\documentclass[10pt,twocolumn,letterpaper]{article}

\usepackage[pagenumbers]{cvpr} 

\usepackage[dvipsnames]{xcolor}

\usepackage[accsupp]{axessibility}
\usepackage{algorithm}
\usepackage{algorithmic}

\usepackage{amsmath}
\DeclareMathOperator*{\argmax}{arg\,max}

\usepackage{multirow}
\usepackage{graphicx}
\usepackage{makecell}
\usepackage{bbding}
\usepackage{xcolor}
\usepackage{diagbox}

\definecolor{cvprblue}{rgb}{0.21,0.49,0.74}
\usepackage[pagebackref,breaklinks,colorlinks,citecolor=cvprblue]{hyperref}

\def\paperID{11832} 
\def\confName{CVPR}
\def\confYear{2024}

\title{GenesisTex: Adapting Image Denoising Diffusion to Texture Space}

\author{
    Chenjian Gao\textsuperscript{1,2\dag\#}
    \quad Boyan Jiang\textsuperscript{2}
    \quad\enspace Xinghui Li\textsuperscript{2}
    \quad Yingpeng Zhang\textsuperscript{2\#}
    \quad\enspace Qian Yu\textsuperscript{1*}
    \\\textsuperscript{1}School of Software, Beihang University \\\textsuperscript{2}\textbf{R\&D Efficiency and Capability Department, Tencent IEG} \\
    \url{https://cjeen.github.io/GenesisTexPaper/}
}
\begin{document}
\maketitle
{\let\thefootnote\relax\footnote{{\textsuperscript{\dag}Work was done during an internship at Tencent IEG.}}}
{\let\thefootnote\relax\footnote{{\textsuperscript{\#}Equal contribution.}}}
{\let\thefootnote\relax\footnote{{\textsuperscript{*}Corresponding author.}}}
\begin{abstract}
We present GenesisTex, a novel method for synthesizing textures for 3D geometries from text descriptions. GenesisTex adapts the pretrained image diffusion model to texture space by texture space sampling. Specifically, we maintain a latent texture map for each viewpoint, which is updated with predicted noise on the rendering of the corresponding viewpoint. The sampled latent texture maps are then decoded into a final texture map. During the sampling process, we focus on both global and local consistency across multiple viewpoints: global consistency is achieved through the integration of style consistency mechanisms within the noise prediction network, and low-level consistency is achieved by dynamically aligning latent textures.  Finally, we apply reference-based inpainting and img2img on denser views for texture refinement. Our approach overcomes the limitations of slow optimization in distillation-based methods and instability in inpainting-based methods. Experiments on meshes from various sources demonstrate that our method surpasses the baseline methods quantitatively and qualitatively.
\end{abstract}    
\section{Introduction}
In recent years, with the development of deep learning, 3D content generation technology has made significant progress. The applications of 3D content generation are diverse, ranging from AR/VR to gaming and filmmaking. While there has been considerable research on deep learning-based geometric asset generation\cite{Sun2024}, there has been a notable industry demand for generating realistic textures for given geometries.

Recently, text-conditioned image diffusion models~\cite{rombach2022high} have achieved impressive results in image generation. Some works~\cite{poole2022dreamfusion, lin2023magic3d, chen2023fantasia3d} have leveraged text-conditioned image diffusion models to generate textured 3D assets by generating content from multiple viewpoints, achieving notable performance. 
In this work, our focus is on generating high-quality textures for a given geometry, leveraging the priors provided by pre-trained text-to-image diffusion models. This task poses several challenges, including: 1) View consistency: ensuring cross-view constraints for maintaining low-level consistency; 2) High efficiency: generating textures for a model within a few minutes, enabling practical applications; 3) Zero-shot learning: achieving texture generation without requiring additional training or finetuning. Addressing these challenges is crucial for successful application of image diffusion models to the texture domain.
\begin{figure}
  \centering
  \includegraphics[width=\linewidth]{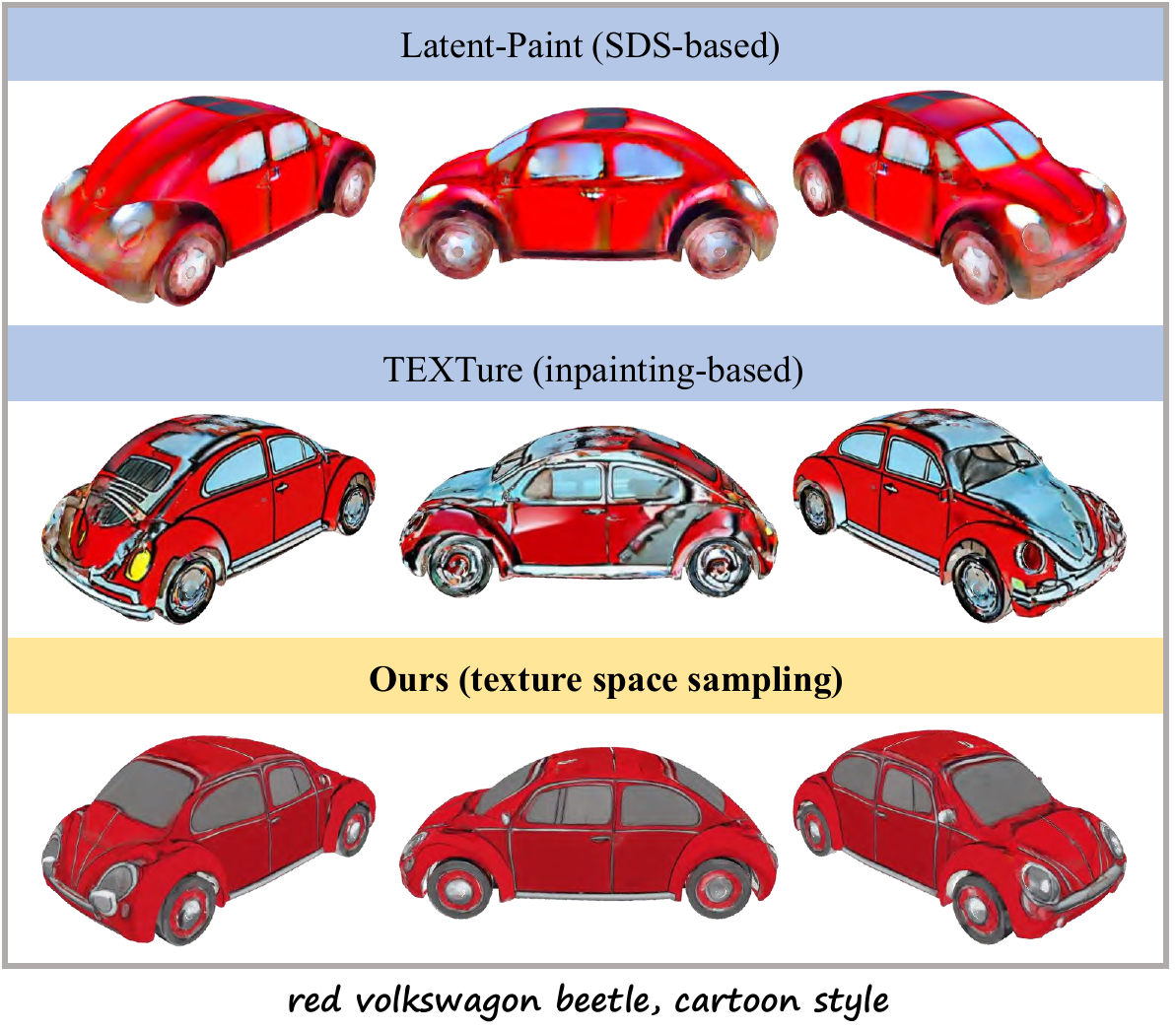}
  \caption{
    Texturing results of different methods. Score Distillation Sampling (SDS) based method produces blurred and oversaturated textures. Inpainting-based approach results in artifacts at the boundaries of inpainting masks. Texture space sampling concurrently generating content from multiple viewpoints, produces clean, clear and natural colored textures.
  }
  \label{fig:intro}
\end{figure}

Currently, the most prominent method for text-to-3D generation is Score Distillation Sampling (SDS)~\cite{poole2022dreamfusion, lin2023magic3d, chen2023fantasia3d}. SDS utilizes the prior knowledge from image diffusion models to iteratively generate gradients for rendering images of 3D objects. However, SDS has limitations in terms of both efficiency and quality. It often takes several hours to generate a single textured 3D object, and the resulting textures can suffer from  over-saturation due to the adoption of a large CFG (classifier-free guidance) scale, which reduces randomness during generation.
Recent works~\cite{chen2023text2tex, richardson2023texture, cao2023texfusion} have introduced inpainting-based methods for multi-view generation. These methods establish a predetermined order of views, using the content of previously generated views as a condition for subsequent views. While inpainting-based approaches are faster and produce realistic colors, they are sensitive to the predefined view order. Additionally, they require texture segmentation to determine the source view for each texture pixel, which often results in artifacts around the borders of the segmented areas.

To address the aforementioned issues, we propose GenesisTex, a novel approach that introduces texture space sampling. Specifically, we maintain multiple-view latent texture maps throughout the sampling process and perform denoising to improve their quality. Subsequently, the content is decoded from the latent space to obtain RGB textures.
Our approach focuses on achieving two aspects of consistency during the sampling process. Firstly, we ensure global style consistency across multiple views by incorporating style consistency in the noise prediction network. This helps maintain a coherent style throughout the generated textures. Secondly, we employ dynamic alignment of latent textures to ensure low-level multi-view consistency, enhancing the overall quality of the generated textures.
Due to memory limitations, texture space sampling operates on sparse views. However, to further enhance the quality, we leverage reference-based inpainting and Img2Img techniques on denser views for texture map refinement.
GenesisTex can generate detailed, clean, and naturally colored textures for a given geometry within a few minutes. 
Unlike inpainting-based methods, our proposed texture space sampling concurrently generates content for multiple views without relying on a predefined order. This adaptability makes GenesisTex suitable for various geometries and results in fewer artifacts in the generated textures.

Our contributions can be summarized as follows:
\begin{itemize}
    \item First, we present a novel method for texture generation, where the core is texture space sampling.  This sampling technique allows for the concurrent denoising of latent textures associated with multiple viewpoints.
    \item Second, we introduce Style Consistency and Dynamic Alignment in texture space sampling for multi-view consistency.
    \item Third, we conduct a comprehensive study involving numerous 3D objects from various sources. The experimental results demonstrate the superiority of our method over baseline methods.
\end{itemize}

\section{Related Work}
\noindent\textbf{Texture Synthesis.}
Generating textures over 3D surfaces is a challenging problem, as it requires attention to both colors and geometry. Earlier works like AUV-Net~\cite{chen2022auv} and Texturify~\cite{siddiqui2022texturify} embed the geometric prior into a UV map or mesh parameterization. Different from them, EG3D~\cite{chan2022efficient} and GET3D~\cite{gao2022get3d} directly train 3D StyleGANs~\cite{karras2019style} to generate geometries and textures jointly, where the textures are implicit texture fields. 
However, these methods either only work on a single category, or demand textured 3D shapes for training without text conditioning, which limits their broad applicability. 
Recently, \cite{chen2023text2tex, richardson2023texture, cao2023texfusion} use the priors provided by the image diffusion model to synthesize textures.
Text2Tex~\cite{chen2023text2tex} and TEXTure~\cite{richardson2023texture} performs inpainting on multiview renderings. TexFusion~\cite{cao2023texfusion} propose a sequential texture sampling method. Our method predicts multi-view noise concurrently, avoiding the following issues of sequential noise prediction: 1) Each iteration requires adding forward noise to visited regions to match unvisited areas, potentially leading to detail loss. 2) Consistency is only constrained between adjacent viewpoints, with no direct consistency for long-range viewpoints.

\noindent\textbf{Text-to-Image Diffusion Models.}
Over the past years, the development of several large-scale diffusion models~\cite{nichol2021glide, ramesh2022hierarchical, rombach2022high, saharia2022photorealistic} has enabled the production of highly detailed and visually impressive images. These models generate images based on input text prompts. Specifically, Stable Diffusion is trained on a substantial text-image dataset~\cite{schuhmann2022laion} and incorporates a text encoder from CLIP~\cite{radford2021learning} to understand the input prompts.
Beyond the basic text conditioning, ControlNet~\cite{zhang2023adding} enables the model to condition its denoising network on additional input modalities, such as depth maps. 
In this work, we utilize ControlNet and Stable Diffusion to provide geometrically-conditioned image priors.

\noindent\textbf{Text-to-3D using 2D Image Diffusion Models.}
Early works~\cite{jain2022zero, mohammad2022clip, michel2022text2mesh} utilize the pretrained CLIP model to maximize the similarity between rendered images and text prompt. However, pioneering works DreamFusion~\cite{poole2022dreamfusion} and SJC~\cite{wang2023score}, on the other hand, propose to distill a 2D text-to-image generation model to generate 3D shapes from texts, and many follow-up works~\cite{lin2023magic3d, liu2023one, liu2023zero, long2022sparseneus, qian2023magic123, wang2023prolificdreamer, metzer2023latent} follow such per-shape optimization scheme. Recently, several methods~\cite{chan2023generative, gu2023nerfdiff, szymanowicz2023viewset, tang2023mvdiffusion, tseng2023consistent, shi2023mvdream, xiang20233d, liu2023syncdreamer, long2023wonder3d, huang2023epidiff} have been proposed to generate consistent multi-view images by using diffusion models. MVDream~\cite{shi2023mvdream} and SyncDreamer~\cite{liu2023syncdreamer} share similar ideas, generating consistent multi-view images via attention layers.
However, existing approaches either endure slow optimization processes or depend on separately trained 3D priors, rendering them unsuitable for direct application in texture synthesis. In contrast, our method does not require additional training and can generate results within several minutes.
\section{Methodology}
Our objective is to generate a texture map $\mathcal{T}$ for a given 3D mesh $\mathcal{M}$, using the provided text as a descriptive condition. Our approach leverages the Stable Diffusion model as the image diffusion model. We begin by introducing some basic concept about Stable Diffusion and texture representation in Sec.~\ref{sec:pre}. Then, we introduce a sampling algorithm in texture space for Stable Diffusion in Sec.~\ref{sec:sampling}. Furthermore, we propose multi-view consistency strategy in Sec.~\ref{sec:consistency}. Finally we perform refinement on the texture map, which includes Inpainting and Img2Img, as described in Sec.~\ref{sec:refine}.
\subsection{Preliminary}
\label{sec:pre}

\noindent \textbf{Stable Diffusion.} The diffusion model belongs to a class of generative models that generate data through iterative denoising from random noise. In the sampling process, $\textbf{z}_{i-1}$ can be sampled from $\textbf{z}_i$ using a DDIM sampler~\cite{song2020denoising}:
\begin{equation}
\begin{aligned}
  \hat{\textbf{z}}_{i\rightarrow0}&=(\textbf{z}_i-\sqrt{1-\alpha_{i}}\epsilon_\theta(\textbf{z}_i, t_i, \mathbf{c}_{text}))/\sqrt{\alpha_{i}},\\
  \textbf{z}_{i-1}=&\sqrt{\alpha_{i-1}}\hat{\textbf{z}}_{i\rightarrow0}
  +\sqrt{1-\alpha_{i-1}-\sigma_{i}^2}\epsilon_\theta(\textbf{z}_i, t_i, \mathbf{c}_{text})\\
  &+ \sigma_{i} \varepsilon,
\end{aligned}
\end{equation}
where $\epsilon_\theta$ is a pretrained U-Net for noise prediction and the initial $\textbf{z}_T\sim\mathcal{N}(0, \mathbf{I})$. In this paper, we specifically employ Stable Diffusion~\cite{rombach2022high}.
Stable Diffusion performs denoising in the latent space and employs an autoencoder $\mathcal{D}(\mathcal{E}(\cdot))$ for the conversion between image and latent representations. So $\textbf{z}_0$ generated by the diffusion process is decoded to the image $\mathcal{D}(\textbf{z}_0)$ finally.

\noindent \textbf{ControlNet.} ControlNet~\cite{zhang2023adding} injects low-level control during the denoising process of Stable Diffusion.
Our approach employs depth $\textbf{d}$ as the geometric control condition. The predicted noise of the U-Net with ControlNet is represented as $\epsilon_\theta(\textbf{z}_i, t_i, \mathbf{c}_{text}, \textbf{d})$. 

\noindent \textbf{Texture Representation.} 
In this paper, we utilize mesh as the 3D representation. Texture map is associated with an UV parameterization of the mesh $\mathcal{M}$. We use \textit{xatlas}~\cite{xatlas2016} to generate the UV parameterization.

\noindent \textbf{Rendering.} Given a mesh $\mathcal{M}$, a texture map $\mathcal{T}$ and a viewpoint $C$, we can use the rendering function $\mathcal{R}$ to get the rendered image $\textbf{x}^{(img)} = \mathcal{R}(\mathcal{T}; \mathcal{M}, C)$. 
Similar to \cite{cao2023texfusion}, we do not consider any lighting and rendering involves solely sampling colors from the texture map. The inverse rendering function $\mathcal{R}^{-1}$ can inverse render the image $\textbf{x}^{(img)}$ back to a texture map $\mathcal{T}^{\prime} = \mathcal{R}^{-1}(\textbf{x}^{(img)}; \mathcal{M}, C)$.

\subsection{Sampling in Texture Space}
\label{sec:sampling}
\begin{figure*}[t]
  \centering
  \includegraphics[width=\linewidth]{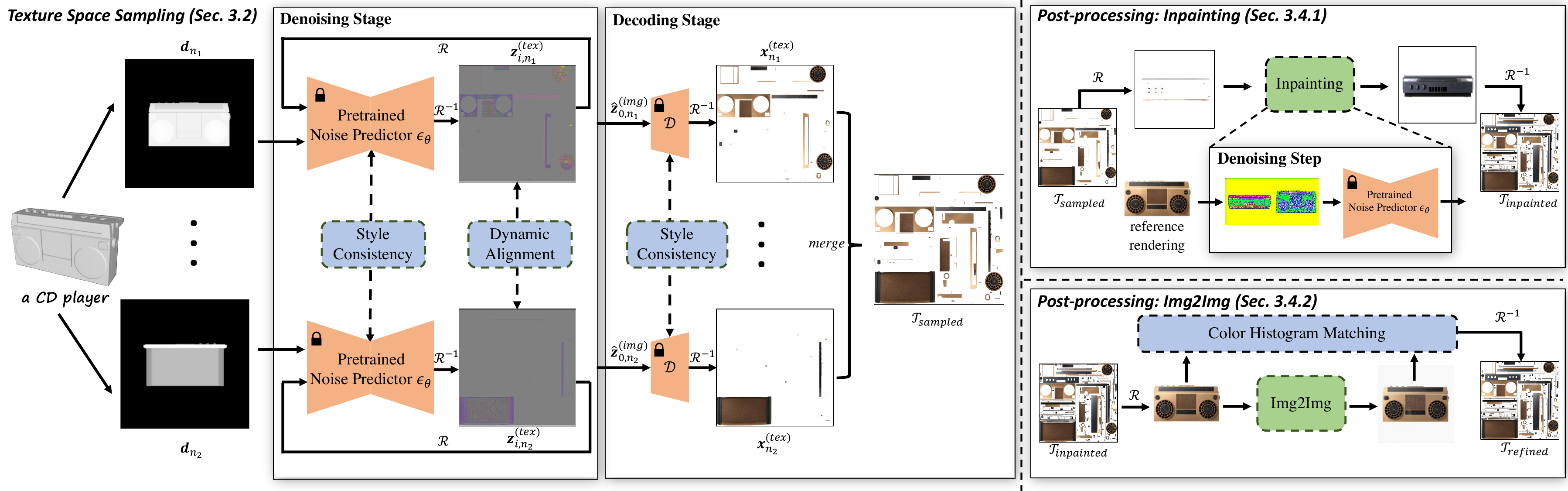}
  \caption{
    Overview of GenesisTex. 
GenesisTex generates a texture map for a given mesh $\mathcal{M}$, based on a prompt. Texture Space Sampling samples a texture map using Stable Diffusion, introducing style consistency and dynamic alignment across multiple viewpoints. Furthermore, Inpainting and Img2Img are applied to fill in the blank regions and enhance the quality of texture map details, respectively.
  }
  \label{fig:architecture}
\end{figure*}
The most straightforward method for sampling on the texture map is to fine-tune the Stable Diffusion model directly on some texture maps, making its distribution more like the distribution of texture maps. However, the distribution of texture maps is much more complex than that of rendering images since different UV parameterizations of the same 3D object correspond to different texture maps. Instead of fine-tuning existing image diffusion models, we adapt them to the texture space for texture map generation.

Algorithm~\ref{alg:texture_sampling} presents our proposed texture space sampling algorithm. Firstly we define a set of $\mathcal{C}^{(sampling)} = \{C_1, .., C_N\}$ camera views and render the corresponding depth map $\textbf{d}_n$ in each view. Since we employ a latent diffusion model, our texture space sampling method first perform denoising in the latent texture space and then decode the latent to texture image. In the denoising stage, a dedicated diffusion denoising algorithm is applied to a set of latent texture maps $\{\mathbf{z}_{n}^{(tex)}\}_{n=1}^N$. In the decoding stage, we recover the texture map $\mathcal{T}_{sampled}$ from the latent space.
\begin{algorithm}[t]
\caption{Texture Space Sampling}\label{alg:texture_sampling}
\begin{algorithmic}
\STATE Input: mesh $\mathcal{M}$, cameras $\{C_1, \dots, C_N \}$
\STATE Parameters: Denoising time schedule $\{t_i\}_{i=T}^0$, DDIM noise schedule $\{\sigma_i\}_{i=T}^0$
\STATE $\{\mathbf{z}_{T, n}^{(tex)}\}_{n=1}^N \leftarrow \{\mathbf{0}\}$
\STATE $\{\mathbf{z}_{n}^{(bg)}\}_{n=1}^N\sim \{\mathcal{N}(\mathbf{0}, \mathbf{I})\}$
\STATE $\{\mathbf{\hat{\epsilon}}_{T+1, n}\}_{n=1}^N\sim \{\mathcal{N}(\mathbf{0}, \mathbf{I})\}$
\STATE \COMMENT \textbf{denoising stage:}
\FOR{$i \in \{T \dots 1\}$}
    \FOR{$n \in \{1 \dots N\}$}
        \STATE $\mathbf{\varepsilon}_{n} \sim \mathcal{N}(\mathrm{0}, \boldsymbol{I})$
        \STATE ${\mathbf{z}_{i, n}^{(img)} \leftarrow \mathbf{M}_n (\sqrt{\alpha_{i}}\mathcal{R}(\mathbf{z}_{i, n}^{(tex)}) + \sqrt{1-\alpha_{i}-\sigma_{i+1}^2} \mathbf{\hat{\epsilon}}_{i+1, n}) }$\\$\quad\quad\quad\quad + (1-\mathbf{M}_n) \mathbf{z}_{n}^{(bg)} + \sigma_{i+1} \varepsilon_{n}$\;
        \STATE $\mathbf{\hat{\epsilon}}_{i, n} \leftarrow \epsilon_\theta(\mathbf{z}_{i, n}^{(img)}, t_{i}, \textbf{c}, \textbf{d}_n)$
        \STATE $\mathbf{\hat{z}}_{0, n}^{(img)} \leftarrow (\mathbf{z}_{i, n}^{(img)}-\sqrt{1-\alpha_{i}}\mathbf{\hat{\epsilon}}_{i, n})/\sqrt{\alpha_{i}}$
        \STATE $\mathbf{z}_{i-1, n}^{(tex)} \leftarrow \mathcal{R}^{-1}(\mathbf{\hat{z}}_{0, n}^{(img)})$
        \STATE $\mathbf{z}_{n}^{(bg)} \leftarrow \sqrt{\alpha_{i-1}} \mathbf{\hat{z}}_{0, n}^{(img)} + \sqrt{1 - \alpha_{i-1} - \sigma_{i}^2} \cdot \mathbf{\hat{\epsilon}}_{i, n}$
    \ENDFOR
    \STATE $\{\mathbf{z}_{i-1, n}^{(tex)}\}_{n=1}^N = dynamic\_align(\{\mathbf{z}_{i-1, n}^{(tex)}\}_{n=1}^N)$
\ENDFOR
\STATE \COMMENT \textbf{decoding stage:}
\FOR{$n \in \{1 \dots N\}$}
\STATE $\mathbf{x}_{n}^{(img)} = \mathcal{D}(\mathbf{\hat{z}}_{0, n}^{(img)})$
\STATE $\mathbf{x}_{n}^{(tex)} = R^{-1}(\mathbf{x}_{n}^{(img)})$
\ENDFOR
\STATE $\mathcal{T} = merge(\{\mathbf{x}_{n}^{(tex)}\})_{n=1}^N$
\RETURN Texture map $\mathcal{T}_{sampled}$
\end{algorithmic}
\end{algorithm}

\noindent\textbf{Denoising Stage.} In the denoising stage, we adapt the DDIM from the rendering space to the texture space and utilize the same parameters as DDIM, including denoising time schedule $\{t_i\}_{i=T}^0$, DDIM noise schedule $\{\sigma_i\}_{i=T}^0$. For each viewpoint, we maintain a separate latent texture which dynamically interact during the denoising process. We begin with $\{\mathbf{z}_{T, n}^{(tex)}\}_{n=1}^N=\{\mathbf{0}\}$, the zero latent texture map corresponding to $N$ viewpoints. At each denoising step, our goal is to predict $\mathbf{z}_{i-1, n}^{(tex)}$ from $\mathbf{z}_{i, n}^{(tex)}$. We first render the latent texture map $\mathbf{z}_{i, n}^{(tex)}$ to the rendering space using the rendering function $\mathcal{R}$, and then calculate the corresponding latent image $\mathbf{z}_{i, n}^{(img)}$:
\begin{equation}\label{eq:mapping}
\begin{aligned}
  \mathbf{z}_{i, n}^{(img)} &= \mathbf{M}_n (\sqrt{\alpha_{i}}\mathcal{R}(\mathbf{z}_{i, n}^{(tex)}) + \sqrt{1-\alpha_{i}-\sigma_{i+1}^2} \mathbf{\hat{\epsilon}}_{i+1, n}) \\ 
  & + (1-\mathbf{M}_n) \mathbf{z}_{n}^{(bg)} + \sigma_{i+1} \varepsilon_{n}
\end{aligned}
\end{equation}
Here, $\mathbf{M}_n$ represents the foreground mask for viewpoint $C_n$ (downsampled to the same resolution as $\mathbf{z}^{(img)}$), where foreground pixels have a value of $1$, and background pixels have a value of $0$. $\hat{\epsilon}_{i+1, n}$ represents the noise predicted by the noise prediction network at step $i+1 \rightarrow i$ (notably, we define $\hat{\epsilon}_{T+1, n} \sim\mathcal{N}(0, \mathbf{I})$).
$\mathbf{z}_{n}^{(bg)}$ is the background of the latent image, which is set to Gaussian noise at initialization. $\varepsilon_n\sim\mathcal{N}(0, \mathbf{I})$ is a random Gaussian noise.
Subsequently, we predict the corresponding noise $\hat{\epsilon}_{i, n} = \epsilon_\theta(\mathbf{z}_{i, n}^{(img)}, t_{i}, \textbf{c}, \textbf{d}_n)$ for the latent image $\mathbf{z}_{i, n}^{(img)}$ conditioned on the depth map $\textbf{d}_n$. With the predicted noise, we can obtain the predicted $\mathbf{\hat{z}}_{0, n}^{(img)}$:
\begin{equation}
\mathbf{\hat{z}}_{0, n}^{(img)} = (\mathbf{z}_{i, n}^{(img)}-\sqrt{1-\alpha_{i}}\mathbf{\hat{\epsilon}}_{i, n})/\sqrt{\alpha_{i}},
\end{equation}
Finally, we inverse render $\mathbf{\hat{z}}_{0, n}^{(img)}$ to the texture space using $\mathcal{R}^{-1}$ and update $\mathbf{z}_{n}^{(bg)}$:
\begin{equation}
\begin{aligned}
\mathbf{z}_{i-1, n}^{(tex)} &= \mathcal{R}^{-1}(\mathbf{\hat{z}}_{0, n}^{(img)}), \\
\mathbf{z}_{n}^{(bg)} = \sqrt{\alpha_{i-1}} \mathbf{\hat{z}}_{0, n}^{(img)} &+ \sqrt{1 - \alpha_{i-1} - \sigma_{i}^2} \cdot \mathbf{\hat{\epsilon}}_{i, n}
\end{aligned}
\end{equation}
In the denoising stage, we repeat the above process $T-1$ times. After each denoising step, we impose consistency constraints on latent texture ${\mathbf{z}_{i-1, n}^{(tex)}}$, which will be introduced in Sec.~\ref{sec:consistency}. Notably, differing from DDIM where the latent includes noise, the latent texture in our method is an estimate of the noise-free version. This is because $\mathcal{R}$ and $\mathcal{R}^{-1}$ in the denoising process require interpolation of the latent, which may distort the noise.

\noindent\textbf{Decoding Stage.} To recover the texture map $\mathcal{T}_{sampled}$, we first decode each viewpoint's latent separately to obtain multi-view rendering. Subsequently, we inverse render the renderings into the texture space. After obtaining multi-view texture maps, we merge them into a single texture map as the output. Similar to \cite{chen2023text2tex, richardson2023texture, cao2023texfusion}, we aim for each pixel on each texture map to come from the viewpoint  where the corresponding point on the mesh is observed most directly. Therefore, our merge is defined as follows:
\begin{equation}
\mathcal{T}_{sampled} = \sum_{n=1}^{N} \textit{Softmax}(\mathcal{R}^{-1}(\mathbf{N_n})) \times \mathbf{x}_{n}^{(tex)}
\end{equation}
$\mathbf{N_n}$ is the similarity mask at viewpoint $C_{n}$, where each pixel represents the cosine similarity between the normal vectors of the visible faces and the reversed view direction. We use $\textit{Softmax}$ instead of $\textit{Max}$ because $\textit{Softmax}$ can result in more natural transition in the texture maps.

\subsection{Consistency between Latent Texture Maps}
\label{sec:consistency}
\subsubsection{Dynamic Alignment}
During sampling in texture space, we maintain $N$ latent texture maps, now we introduce an alignment approach to ensure their local consistency. Firstly, we directly reduce these latent texture maps to obtain a uniform latent texture map:
\begin{equation}\label{eq:unitex}
\mathbf{z}_{i-1}^{(unitex)}=\sum_{n=1}^{N} \textit{Softmax}(\mathcal{R}^{-1}(\mathbf{N}_n)) \times \mathbf{z}_{i-1, n}^{(tex)}
\end{equation}
Next, we blend each viewpoint's latent texture map $\mathbf{z}_{i-1, n}^{(tex)}$ with the uniform latent texture map $\mathbf{z}_{i-1}^{(unitex)}$.
We do not enforce strict alignment among the $N$ latent texture maps at every step. Since a latent pixel gains full meaning only when combined with the surrounding context, which is viewpoint-dependent, the latent texture map from each viewpoint should contain some independent information. Moreover, the content generated at different timestamps varies, so the alignment constraints should adapt flexibly. We introduce a signal $c(t)$ for dynamically regulating the variations in consistency during the sampling process:
\begin{equation}
\mathbf{z}_{i-1, n}^{(consistex)}= c(t_i) \times \mathbf{z}_{i-1}^{(unitex)} + (1 - c(t_i)) \times \mathbf{z}_{i-1, n}^{(tex)}
\end{equation}
We empirically find that maintaining strong local consistency in the mid-term of denoising, while keeping weak local consistency in the early and late stages, leads to higher quality generation results.

\subsubsection{Style Consistency}
In texture space sampling, we use the noise estimation network $\epsilon_\theta$ to estimate noise for the multi-view latents, which is responsible for content generation. So we adapt $\epsilon_\theta$ to ensure style consistency of estimated noise across different views.
Inspired by video diffusion models \cite{khachatryan2023text2video, yang2023rerender} and multi-view diffusion model \cite{shi2023mvdream}, we modify all the self-attention layers and group normalization layers in the noise prediction network to align the style of multi-view content. 

\noindent\textbf{Adapted Self-Attention.} In Stable Diffusion, 
self-attention facilitates long-range interactions among features within an image. We transform self-attention into cross-view attention to establish style consistency across multiple viewpoints, by using the key $\textbf{K}'$ and value $\textbf{V}'$ from all viewpoints,
\begin{equation}
\begin{aligned}
  \textbf{K}'&=\textbf{W}^K[\mathbf{z}_{i, 1}^{(img)}; ...; \mathbf{z}_{i, n}^{(img)}], \\
  \textbf{V}'&=\textbf{W}^V[\mathbf{z}_{i, 1}^{(img)}; ...; \mathbf{z}_{i, n}^{(img)}].
\end{aligned}
\end{equation}
where $\textbf{W}^K$, $\textbf{W}^V$ are pretrained parameters.

\noindent\textbf{Adapted Group Normalization.} Stable diffusion adopt group normalization as the normalization layer. Group normalization divides the channels into groups and computes within each group the mean and variance for normalization. We convert 2D group normalization to 3D by connecting all different views in the group normalization layer, \ie, the mean and variance are calculated within the channel group of all pixels across all viewpoints.

By adjusting self-attention layer and group normalization layer, we achieve multi-view style consistency without any additional training. In addition to the noise prediction network $\epsilon_\theta$, we apply the same modifications to the decoder $\mathcal{D}$ to further ensure style consistency in the multi-view decoded RGB images. 

\subsection{Texture Map Refinement}
\label{sec:refine}
\subsubsection{Post-processing: Inpainting} 
Since the memory cost of texture space sampling increases with the number of viewpoints of $\mathcal{C}_{sampling}$, the number of viewpoints $N$ is limited. Some regions in the texture map $\mathcal{T}_{sampled}$ may not be observed in any of the viewpoints in $\mathcal{C}_{sampling}$. Therefore, after texture space sampling, we introduce an inpainting epoch to fill texture in areas not observed in $\mathcal{C}$. 

Firstly we define a denser set of viewpoints $\mathcal{C}^{(inpainting)} = \{C_1, .., C_{N_1}\}$ as compared to $\mathcal{C}^{(samping)}$, where $N_1 > N$. We use a mask $\mathbf{M}^{(blank)}_C$ to indicate the rendered areas with blank textures at the viewpoint $C$. 
The entire inpainting epoch comprises $N_1$ iterations. In each iteration, we first compute the viewpoint $C_{n} = \argmax_C \mathbf{M}^{(blank)}_C$ with the largest blank area. Then, we perform inpainting on the rendering $\mathbf{x}_n^{(img)}$ using the pretrained depth conditioned Stable Diffusion model as described in \cite{chen2023text2tex, richardson2023texture}. During inpainting, Stable Diffusion utilizes regions with textures as conditions to fill in blank areas without textures. To enhance the robustness of inpainting, we render an image using a viewpoint $C_0$ from the set $\mathcal{C}^{(sampling)}$. We then concatenate this image to the image requiring inpainting as a additional condition for inpainting. Compared to naive inpainting, reference-based inpainting provides a more comprehensive condition. To ensure a natural and smooth transition between blank areas and textured regions, we apply Gaussian blur to the mask $\mathbf{M}^{(blank)}_{C_{n}}$. 
Finally the inpainted rendering $\mathbf{x}_n^{(img)\prime}$ is used to update the texture map:
\begin{equation}
\begin{aligned}
  \mathcal{T}_{n} &= \mathcal{T}_{n-1} \times (1 - \mathcal{R}^{-1}(\mathbf{M}^{(blank)}) \\&+ \mathcal{R}^{-1}(\mathbf{x}_n^{(img)\prime}) \times \mathcal{R}^{-1}(\mathbf{M}^{(blank)})
\end{aligned}
\end{equation}
We start with$ \mathcal{T}_0 = \mathcal{T}_{sampled}$ and iterate for $N_1$ times. Finally we get $\mathcal{T}_{inpainted} = \mathcal{T}_{N_1}$.

It's worth noting that, unlike \cite{chen2023text2tex, cao2023texfusion, richardson2023texture}, which relies heavily on inpainting to maintain multi-view consistency, in our method inpainting is performed only to patch small areas. Therefore, the inpainting's performance does not significantly affect the final quality of the generated texture.

\subsubsection{Post-processing: Img2Img}
So far our generated texture map $\mathcal{T}_{inpainted}$ may still contain unnatural transitions in some regions due to multi-view conflicts and inpainting artifacts. To further improve the generation quality, we introduce an Img2Img epoch. 

In Img2Img epoch, we also define a set of viewpoints $C^{(img2img)}$ include $N_2$ viewpoints and iterate through viewpoint from $C^{(img2img)}$. In each iteration, we encode the rendering image $\mathbf{x}_n^{(img)}$ to the latent space using the encoder $\mathcal{E}$ and add some noise to the latent rendering. Then we perform denoising using the pretrained depth conditioned Stable Diffusion model. And finally we decode the denoised latent back to image space using $\mathcal{D}$. However, $\mathcal{D}(\mathcal{E}(\cdot))$ is not an identity transformation because of the reconstruction distortion of the autoencoder. So we apply the color histogram matching to the foreground of the decoded image to alleviate the color distortion. Finally we update the texture map using the img2img result $\mathbf{x}_n^{(img)\prime}$ based on view similarity $\mathbf{N}_{n}$ :
\begin{equation}
\begin{aligned}
  \mathcal{T}_{n} &= \mathcal{T}_{n-1} \times (1 - \mathcal{R}^{-1}(\mathbf{N}_{n})) \\&+ \mathcal{R}^{-1}(\mathbf{x}_n^{(img)\prime}) \times \mathcal{R}^{-1}(\mathbf{N}_{n})
\end{aligned}
\end{equation}
We start with$ \mathcal{T}_0 = \mathcal{T}_{inpainted}$. After $N_2$ iterations we get $\mathcal{T}_{refined} = \mathcal{T}_{N_2}$.

\section{Experiments}
\subsection{Setup}
\textbf{Implementation Details.} Our experiments are conducted on an NVIDIA A10 GPU. We utilize the official Stable Diffusion-v1.5 model with the ControlNet-v1.1 (depth). For Inpainting epoch and Img2Img epoch, we use DDIM~\cite{song2020denoising} as the sampler. For all samplers, we set the number of iterations to 20 steps and the CFG scale (classifier-free guidance scale) to 7.5. We implement the rendering function using nvdiffrast~\cite{laine2020modular} and make modifications to nvdiffrast to support inverse rendering. Texture space sampling takes about 3 minutes to generate textures for a single object. For more detailed parameter settings, please refer to the supplementary materials.
\begin{figure*}
  \centering
  \includegraphics[width=\linewidth]{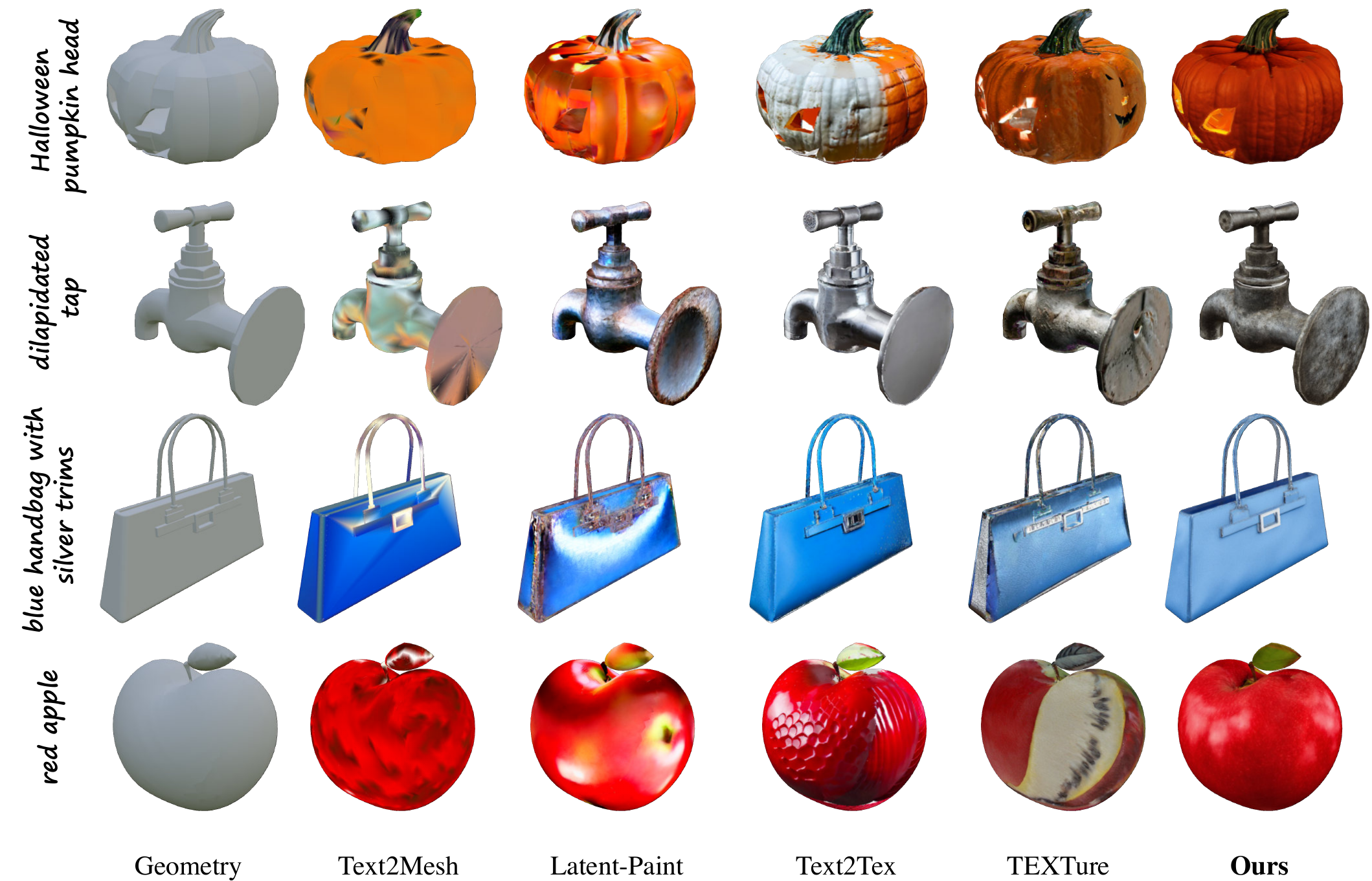}
  \caption{
    Qualitative comparisons with Text2Mesh~\cite{michel2022text2mesh}, Latent-Paint~\cite{metzer2023latent}, Text2Tex~\cite{chen2023text2tex} and TEXTure~\cite{richardson2023texture}. In comparison with the baselines, our GenesisTex exhibits richer details and fewer artifacts.
  }
  \label{fig:compare}
\end{figure*}

\begin{figure}
  \centering
  \includegraphics[width=\linewidth]{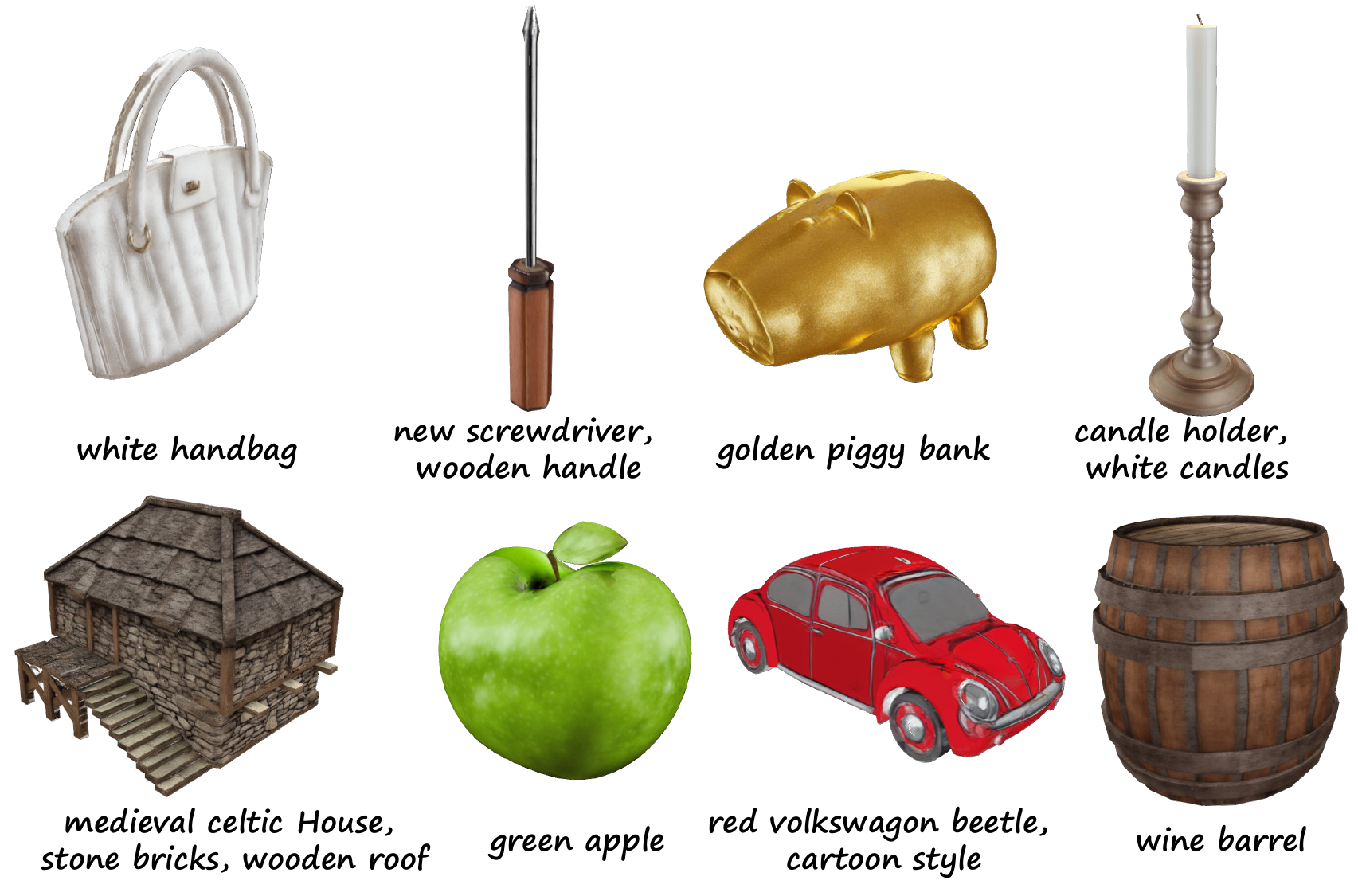}
  \caption{
    Texturing results with GenesisTex.
  }
  \label{fig:examples}
\end{figure}

\begin{figure}
  \centering
  \includegraphics[width=\linewidth]{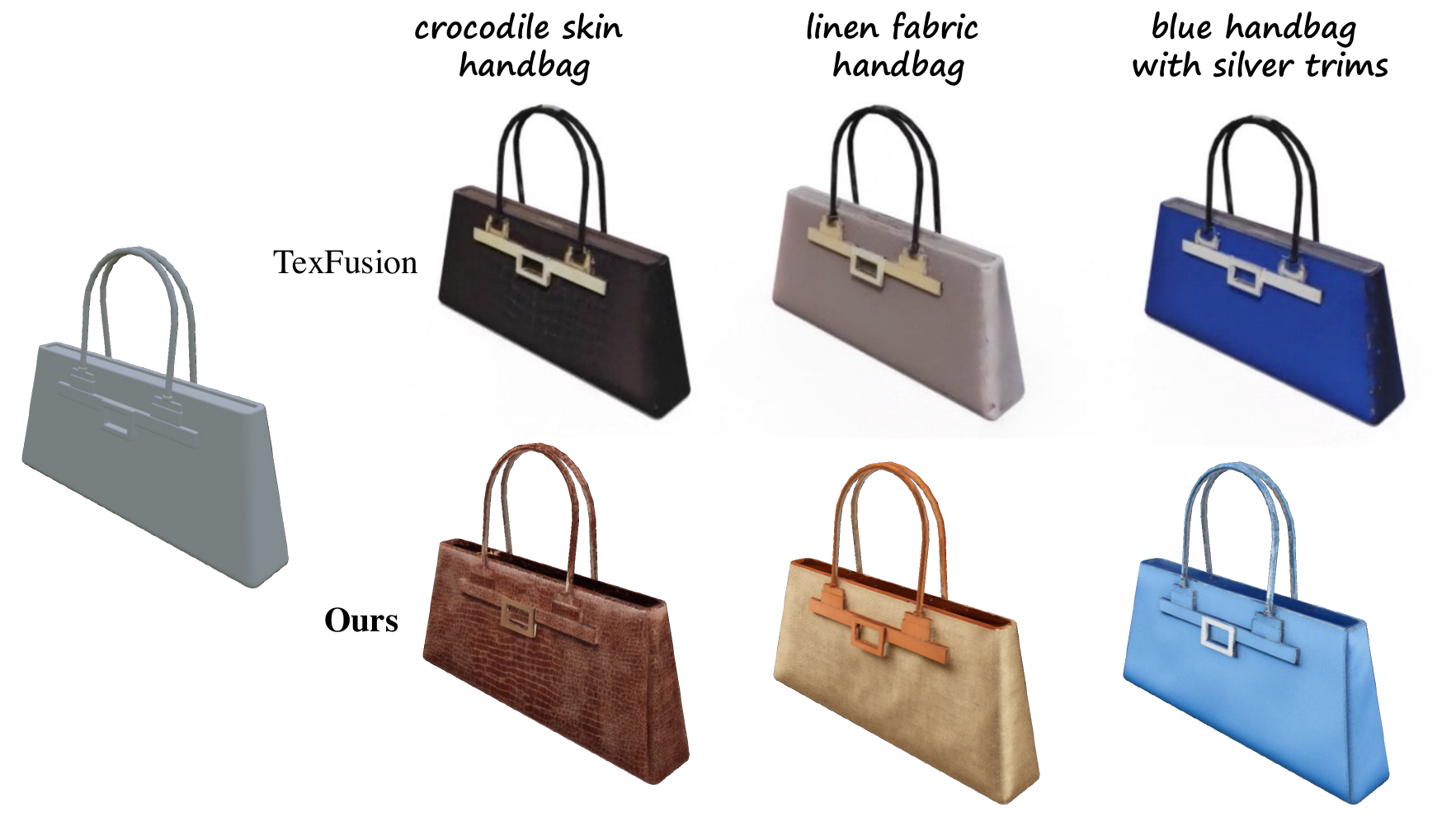}
  \caption{
    Qualitative Comparisons with TexFusion~\cite{cao2023texfusion}. Images are extracted from their original paper.
  }
  \label{fig:compare_texfusion}
\end{figure}

\noindent\textbf{Dataset.} Similar to TexFusion~\cite{cao2023texfusion}, we utilize 35 meshes to test the texture generation performance, including 11 objects from Objaverse~\cite{deitke2023objaverse}, 17 from ShapeNet~\cite{chang2015shapenet}, 1 from Text2Mesh~\cite{michel2022text2mesh}, 2 from Turbosquid~\cite{turbosquid}, 2 from Stanford 3D Scans~\cite{turk1994zippered, curless1996volumetric} and 2 from Three D Scans~\cite{3dscans}. Each object has 1-4 text descriptions, resulting in a total of 80 (mesh, prompt) pairs in this collection.

\noindent\textbf{Baselines.} We conduct comparisons with state-of-the-art methods for generating textures from text: 1) \textbf{Text2Mesh}~\cite{michel2022text2mesh}, a method that stylizes a 3D mesh by predicting color and local geometric details which conform to a target text prompt, harnessing the representational power of CLIP. 2) \textbf{Latent-Paint}~\cite{metzer2023latent}, an approach leveraging SDS to obtain texture gradients from the Stable Diffusion model. 3) \textbf{Text2Tex}~\cite{chen2023text2tex}, a method that uses a depth-aware image inpainting diffusion model to incrementally produce partial textures from various viewpoints. 4) \textbf{TEXTure}~\cite{richardson2023texture}, similar to Text2Tex, but utilizes a different region segmentation strategy. 5) \textbf{TexFusion}~\cite{cao2023texfusion}, a method sequentially inpainting latent texture. As TexFusion does not have publicly available source code, we extract some rendered images of generated results from their paper for comparison.

\subsection{Qualitative Analysis}
Fig.~\ref{fig:compare} and Fig.~\ref{fig:compare_texfusion} presents visual comparisons between our method and baseline approaches. Textures produced by Text2Mesh are notably lacking in detail. Latent-Paint yields textures with greater intricacy compared to Text2Mesh, but they still exhibit considerable blurriness due to SDS limitations. Inpainting-based approaches like Text2Tex and TEXTure introduce artifacts with unnatural transitions at inpainting mask boundaries, struggling to maintain consistency across multiple views because they depend heavily on inpainting. The output of TexFusion is compromised by blurry details, likely a consequence of repeated noise addition during sequential sampling. In contrast, our method stands out by producing textures that are not only clear but also remarkably coherent.

Fig.~\ref{fig:examples} presents more results of our method. Benefiting from the powerful prior of Stable Diffusion, our method can generate textures of diverse styles for different geometries.

\subsection{Quantitative Comparisons}
\begin{table}
\small
\centering
\resizebox{0.49\textwidth}{!}{
\begin{tabular}{l c c c c} 
\toprule
\multirow{2}{*}{Method} & \multirow{2}{*}{FID ($\downarrow$)}  & KID ($\downarrow$) & \multicolumn{2}{c}{User study (\%)} \\
 & &($\times10^{-3}$) &  \makecell[c]{Visual\\ Quality ($\uparrow$)} & \makecell[c]{Align with\\ Prompt ($\uparrow$)}  \\
\midrule
Latent-Paint   & 110.14 & 10.64 & 5.15 & 4.28 \\
Text2Mesh   & 121.61 & 15.13 & 3.15 & 4.86 \\
Text2Tex   & 101.38 & 8.35 & 18.28 & 19.72 \\
TEXTure    & 100.47 & 9.22 & 23.42 & 24.86 \\
$\textbf{Ours}$ & $\textbf{74.58}$ & $\textbf{2.89}$  & $\textbf{50.00}$ & $\textbf{46.28}$ \\
\bottomrule
\end{tabular}
}
\caption{ \small Quantitative comparisons with baseline methods.
}
\label{tab: comp}
\end{table}
\begin{table}
\small
\centering
\resizebox{0.49\textwidth}{!}{
\begin{tabular}{c c c | c c} 
\toprule
\makecell[c]{texture space \\sampling} & \makecell[c]{Inpainting\\Round} & \makecell[c]{Img2Img\\Round} & FID($\downarrow$) & \makecell[c]{KID($\downarrow$)\\($\times10^{-3}$)} \\
\midrule
\textcolor{olive}{\CheckmarkBold} & \textcolor{violet}{\XSolidBrush} & \textcolor{violet}{\XSolidBrush}  & 86.25 & 3.79\\
\textcolor{olive}{\CheckmarkBold}& \textcolor{violet}{\XSolidBrush} & \textcolor{olive}{\CheckmarkBold} & 82.05 & 3.67\\
\textcolor{olive}{\CheckmarkBold}& \textcolor{olive}{\CheckmarkBold}& \textcolor{violet}{\XSolidBrush} & 76.30 & 3.29\\
\textcolor{olive}{\CheckmarkBold}& \textcolor{olive}{\CheckmarkBold}& \textcolor{olive}{\CheckmarkBold} & \textbf{74.58} & \textbf{2.89}\\
\bottomrule
\end{tabular}
}
\caption{ \small Effectiveness of texture refinement.
}
\vspace{-0.1in}
\label{tab: abl}
\end{table}

\noindent \textbf{FID \& KID.} Similar to TexFusion, we sample from pretrained image diffusion model to create ground truth labels. 
For each mesh, we generate depth maps from the five most common canonical viewpoints: front, back, top, and both sides. These depth maps, along with textual descriptions, serve as inputs to condition the Stable Diffusion model. To ensure the focus remains on the texture of the objects, we modify the background of the ground truth images to be white. We visualize the textures created by various methods from the same five viewpoints and evaluate their quality by calculating the Fréchet Inception Distance (FID) and Kernel Inception Distance (KID). FID and KID measure the feature dissimilarity between two image collections, with feature extraction performed using the Inception V3~\cite{szegedy2016rethinking}.
The results in Tab.\ref{tab: comp} demonstrates that the textures generated by GenesisTex are closer to the ground truth compared to those produced by baselines, indicating a superior synthesis quality.

\noindent \textbf{User Study.} We conduct a user study to compare the quality of textures generated by our method versus those produced by baseline methods. 
We render all textured mesh results into videos to facilitate the presentation to users. Each video contains a rotating textured mesh along with the corresponding prompt. For each questionnaire, we randomly show the users $10$ groups of rendered videos. 
Each group displays results for a specific (mesh, prompt) pair, comparing outputs from baseline methods and our approach. Each volunteer is required to answer two questions for a single group, including which one has the highest visual quality and which one aligns best with the prompt. Additional information about the questionnaire is available in the supplementary. Finally, we received $35$ valid responses from the questionnaires, which included the results of comparisons from $350$ groups. The results of the user study are shown in Tab.~\ref{tab: comp}. In terms of both visual quality and alignment with the text prompt, our method is preferred by the most participants, with percentages reaching $50.00\%$ and $46.28\%$, respectively. Furthermore, we conduct a pairwise comparison study and the results are included in the supplementary.

\subsection{Ablation Studies}
\label{sec: abl}
\begin{figure}[b]
  \centering
  \includegraphics[width=\linewidth]{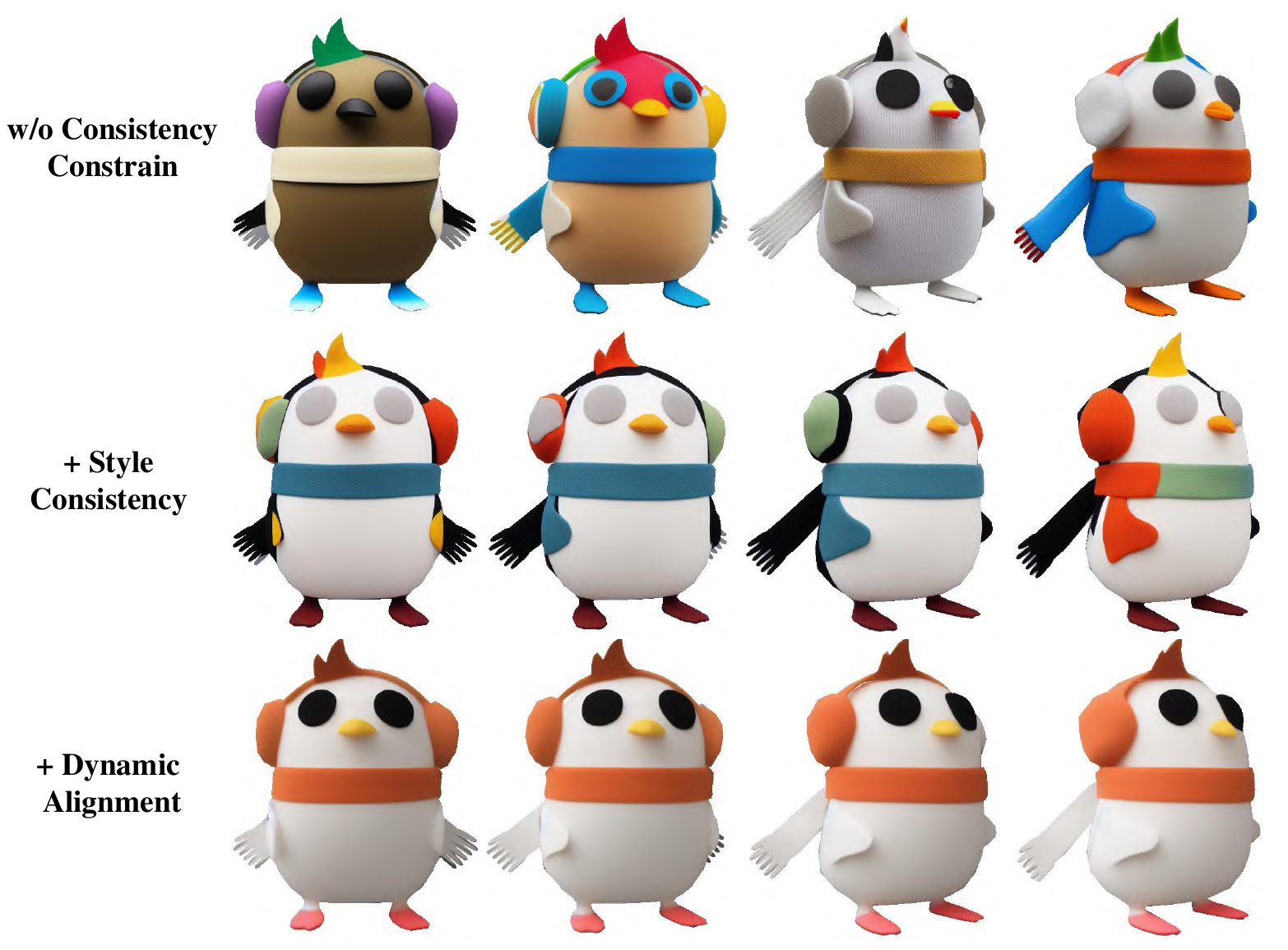}
  \caption{
    Ablation results on consistency in texture space sampling. \texttt{Style Consistency + Dynamic Alignment} (bottom row) achieves the best multi-view consistency.
  }\label{fig:abl_consis}
\end{figure}
\textbf{Consistency in Texture Space Sampling.} Our GenesisTex employs style consistency and dynamic alignment for maintaining multi-view consistency during texture space sampling. To investigate the impact of these two consistency strategies, we visualize the decoded multi-view images. We set $\mathcal{C}(elevation, azimuth) = \{(0^\circ, 0^\circ), (0^\circ, 15^\circ), (0^\circ, 35^\circ), (0^\circ, 45^\circ)\}$ to ensure that multiple viewpoints share sufficient content. Fig.~\ref{fig:abl_consis} illustrates an example with the prompt \textit{penguin toy}. We can observe that without any consistency constraints, the color of the toy varies significant across different viewpoints. After introducing style consistency, consistency for the toy color improves, but there are still inconsistencies in the hair and ears. Further we use dynamic alignment and we can see the consistency of details has significantly improved. We include more results in the supplementary to further demonstrate the effectiveness of these two consistencies.

\begin{figure}[t]
  \centering
  \includegraphics[width=\linewidth]{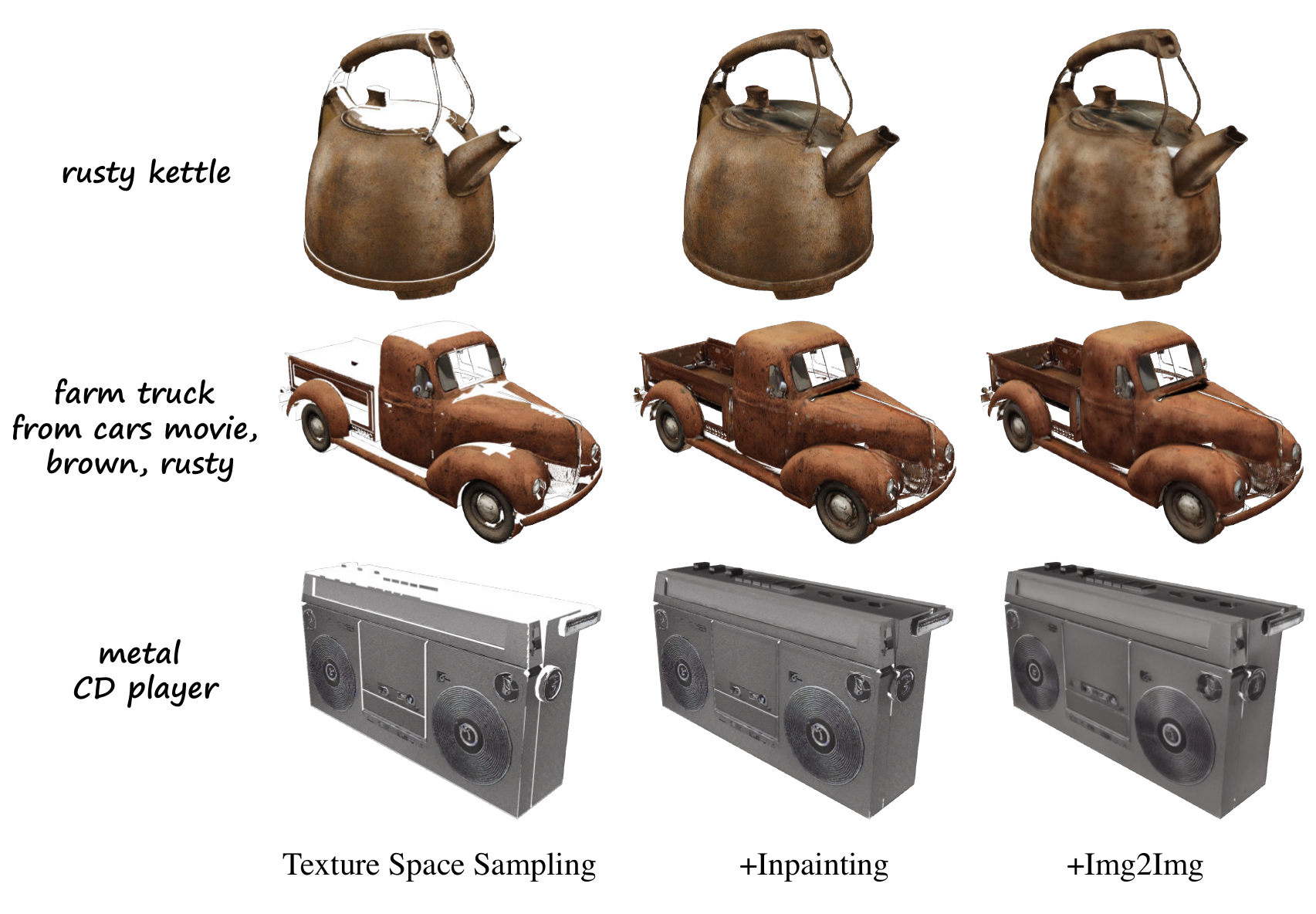}
  \caption{
    Ablation results on texture refinement. Inpainting fills the blank regions, while Img2Img enhances the quality of details.
  }
  \label{fig:abl_comp}
\end{figure}

\noindent\textbf{Effectiveness of Texture Refinement.} After texture space sampling, we conduct refinement on the texture, including Inpainting and Img2Img. We validate the effectiveness of these two modules and the results are presented in Fig.~\ref{fig:abl_comp} and Tab.~\ref{tab: abl}. Inpainting fills in areas lacking texture obtained during texture space sampling, contributing significantly to the improvement in visual quality. Img2Img further repairs artifacts and enhances details, leading to a certain improvement in visual quality as well.

\section{Conclusion and Limitations}
In this work, we propose a novel method for text-based texture generation, named \textit{GenesisTex}. GenesisTex leverages the prior of the pre-trained Stable Diffusion model by introducing texture space sampling. Texture space sampling concurrently generates multi-view content without relying on a predefined sequence of views. Our approach can generate high-quality textures for a given 3D model in several minutes. 
The primary limitation of our method is the significant memory cost associated with maintaining style consistency, caused by cross-view attention. This limitation restricts the number of viewpoints and necessitates post-processing steps, such as inpainting and img2img. Future work could investigate hierarchical style consistency approaches to reduce the computational costs of cross-view attention by iterating over a smaller set of viewpoints.

\noindent \textbf{Acknowledgement}: This work is supported by the Young Elite Scientists Sponsorship Program by CAST (China Association for Science and Technology) and IEG Moonshot Program by Tencent.
{
    \small
    \bibliographystyle{ieeenat_fullname}
    \bibliography{main}

\begin{thebibliography}{54}
\providecommand{\natexlab}[1]{#1}
\providecommand{\url}[1]{\texttt{#1}}
\expandafter\ifx\csname urlstyle\endcsname\relax
  \providecommand{\doi}[1]{doi: #1}\else
  \providecommand{\doi}{doi: \begingroup \urlstyle{rm}\Url}\fi

\bibitem[tur()]{turbosquid}
Turbosquid by shutterstock.
\newblock In \emph{www.turbosquid.com}.

\bibitem[Abid et~al.(2019)Abid, Abdalla, Abid, Khan, Alfozan, and Zou]{abid2019gradio}
Abubakar Abid, Ali Abdalla, Ali Abid, Dawood Khan, Abdulrahman Alfozan, and James Zou.
\newblock Gradio: Hassle-free sharing and testing of ml models in the wild.
\newblock \emph{arXiv preprint arXiv:1906.02569}, 2019.

\bibitem[Cao et~al.(2023)Cao, Kreis, Fidler, Sharp, and Yin]{cao2023texfusion}
Tianshi Cao, Karsten Kreis, Sanja Fidler, Nicholas Sharp, and Kangxue Yin.
\newblock Texfusion: Synthesizing 3d textures with text-guided image diffusion models.
\newblock In \emph{Proceedings of the IEEE/CVF International Conference on Computer Vision}, pages 4169--4181, 2023.

\bibitem[Chan et~al.(2022)Chan, Lin, Chan, Nagano, Pan, De~Mello, Gallo, Guibas, Tremblay, Khamis, et~al.]{chan2022efficient}
Eric~R Chan, Connor~Z Lin, Matthew~A Chan, Koki Nagano, Boxiao Pan, Shalini De~Mello, Orazio Gallo, Leonidas~J Guibas, Jonathan Tremblay, Sameh Khamis, et~al.
\newblock Efficient geometry-aware 3d generative adversarial networks.
\newblock In \emph{Proceedings of the IEEE/CVF Conference on Computer Vision and Pattern Recognition}, pages 16123--16133, 2022.

\bibitem[Chan et~al.(2023)Chan, Nagano, Chan, Bergman, Park, Levy, Aittala, De~Mello, Karras, and Wetzstein]{chan2023generative}
Eric~R Chan, Koki Nagano, Matthew~A Chan, Alexander~W Bergman, Jeong~Joon Park, Axel Levy, Miika Aittala, Shalini De~Mello, Tero Karras, and Gordon Wetzstein.
\newblock Generative novel view synthesis with 3d-aware diffusion models.
\newblock \emph{arXiv preprint arXiv:2304.02602}, 2023.

\bibitem[Chang et~al.(2015)Chang, Funkhouser, Guibas, Hanrahan, Huang, Li, Savarese, Savva, Song, Su, et~al.]{chang2015shapenet}
Angel~X Chang, Thomas Funkhouser, Leonidas Guibas, Pat Hanrahan, Qixing Huang, Zimo Li, Silvio Savarese, Manolis Savva, Shuran Song, Hao Su, et~al.
\newblock Shapenet: An information-rich 3d model repository.
\newblock \emph{arXiv preprint arXiv:1512.03012}, 2015.

\bibitem[Chen et~al.(2023{\natexlab{a}})Chen, Siddiqui, Lee, Tulyakov, and Nie{\ss}ner]{chen2023text2tex}
Dave~Zhenyu Chen, Yawar Siddiqui, Hsin-Ying Lee, Sergey Tulyakov, and Matthias Nie{\ss}ner.
\newblock Text2tex: Text-driven texture synthesis via diffusion models.
\newblock \emph{arXiv preprint arXiv:2303.11396}, 2023{\natexlab{a}}.

\bibitem[Chen et~al.(2023{\natexlab{b}})Chen, Chen, Jiao, and Jia]{chen2023fantasia3d}
Rui Chen, Yongwei Chen, Ningxin Jiao, and Kui Jia.
\newblock Fantasia3d: Disentangling geometry and appearance for high-quality text-to-3d content creation.
\newblock \emph{arXiv preprint arXiv:2303.13873}, 2023{\natexlab{b}}.

\bibitem[Chen et~al.(2022)Chen, Yin, and Fidler]{chen2022auv}
Zhiqin Chen, Kangxue Yin, and Sanja Fidler.
\newblock Auv-net: Learning aligned uv maps for texture transfer and synthesis.
\newblock In \emph{Proceedings of the IEEE/CVF Conference on Computer Vision and Pattern Recognition}, pages 1465--1474, 2022.

\bibitem[Curless and Levoy(1996)]{curless1996volumetric}
Brian Curless and Marc Levoy.
\newblock A volumetric method for building complex models from range images.
\newblock In \emph{Proceedings of the 23rd annual conference on Computer graphics and interactive techniques}, pages 303--312, 1996.

\bibitem[Deitke et~al.(2023)Deitke, Schwenk, Salvador, Weihs, Michel, VanderBilt, Schmidt, Ehsani, Kembhavi, and Farhadi]{deitke2023objaverse}
Matt Deitke, Dustin Schwenk, Jordi Salvador, Luca Weihs, Oscar Michel, Eli VanderBilt, Ludwig Schmidt, Kiana Ehsani, Aniruddha Kembhavi, and Ali Farhadi.
\newblock Objaverse: A universe of annotated 3d objects.
\newblock In \emph{Proceedings of the IEEE/CVF Conference on Computer Vision and Pattern Recognition}, pages 13142--13153, 2023.

\bibitem[Gao et~al.(2022)Gao, Shen, Wang, Chen, Yin, Li, Litany, Gojcic, and Fidler]{gao2022get3d}
Jun Gao, Tianchang Shen, Zian Wang, Wenzheng Chen, Kangxue Yin, Daiqing Li, Or Litany, Zan Gojcic, and Sanja Fidler.
\newblock Get3d: A generative model of high quality 3d textured shapes learned from images.
\newblock \emph{Advances In Neural Information Processing Systems}, 35:\penalty0 31841--31854, 2022.

\bibitem[Gu et~al.(2023)Gu, Trevithick, Lin, Susskind, Theobalt, Liu, and Ramamoorthi]{gu2023nerfdiff}
Jiatao Gu, Alex Trevithick, Kai-En Lin, Joshua~M Susskind, Christian Theobalt, Lingjie Liu, and Ravi Ramamoorthi.
\newblock Nerfdiff: Single-image view synthesis with nerf-guided distillation from 3d-aware diffusion.
\newblock In \emph{International Conference on Machine Learning}, pages 11808--11826. PMLR, 2023.

\bibitem[Huang et~al.(2023)Huang, Wen, Dong, Wang, Li, Chen, Cao, Liang, Qiao, Dai, and Sheng]{huang2023epidiff}
Zehuan Huang, Hao Wen, Junting Dong, Yaohui Wang, Yangguang Li, Xinyuan Chen, Yan-Pei Cao, Ding Liang, Yu Qiao, Bo Dai, and Lu Sheng.
\newblock Epidiff: Enhancing multi-view synthesis via localized epipolar-constrained diffusion, 2023.

\bibitem[Jain et~al.(2022)Jain, Mildenhall, Barron, Abbeel, and Poole]{jain2022zero}
Ajay Jain, Ben Mildenhall, Jonathan~T Barron, Pieter Abbeel, and Ben Poole.
\newblock Zero-shot text-guided object generation with dream fields.
\newblock In \emph{Proceedings of the IEEE/CVF Conference on Computer Vision and Pattern Recognition}, pages 867--876, 2022.

\bibitem[Karras et~al.(2019)Karras, Laine, and Aila]{karras2019style}
Tero Karras, Samuli Laine, and Timo Aila.
\newblock A style-based generator architecture for generative adversarial networks.
\newblock In \emph{Proceedings of the IEEE/CVF conference on computer vision and pattern recognition}, pages 4401--4410, 2019.

\bibitem[Khachatryan et~al.(2023)Khachatryan, Movsisyan, Tadevosyan, Henschel, Wang, Navasardyan, and Shi]{khachatryan2023text2video}
Levon Khachatryan, Andranik Movsisyan, Vahram Tadevosyan, Roberto Henschel, Zhangyang Wang, Shant Navasardyan, and Humphrey Shi.
\newblock Text2video-zero: Text-to-image diffusion models are zero-shot video generators.
\newblock \emph{arXiv preprint arXiv:2303.13439}, 2023.

\bibitem[Laine et~al.(2020)Laine, Hellsten, Karras, Seol, Lehtinen, and Aila]{laine2020modular}
Samuli Laine, Janne Hellsten, Tero Karras, Yeongho Seol, Jaakko Lehtinen, and Timo Aila.
\newblock Modular primitives for high-performance differentiable rendering.
\newblock \emph{ACM Transactions on Graphics (TOG)}, 39\penalty0 (6):\penalty0 1--14, 2020.

\bibitem[Laric(2012)]{3dscans}
Oliver Laric.
\newblock Three d scans.
\newblock In \emph{threedscans.com}, 2012.

\bibitem[Lin et~al.(2023)Lin, Gao, Tang, Takikawa, Zeng, Huang, Kreis, Fidler, Liu, and Lin]{lin2023magic3d}
Chen-Hsuan Lin, Jun Gao, Luming Tang, Towaki Takikawa, Xiaohui Zeng, Xun Huang, Karsten Kreis, Sanja Fidler, Ming-Yu Liu, and Tsung-Yi Lin.
\newblock Magic3d: High-resolution text-to-3d content creation.
\newblock In \emph{Proceedings of the IEEE/CVF Conference on Computer Vision and Pattern Recognition}, pages 300--309, 2023.

\bibitem[Liu et~al.(2023{\natexlab{a}})Liu, Xu, Jin, Chen, Xu, Su, et~al.]{liu2023one}
Minghua Liu, Chao Xu, Haian Jin, Linghao Chen, Zexiang Xu, Hao Su, et~al.
\newblock One-2-3-45: Any single image to 3d mesh in 45 seconds without per-shape optimization.
\newblock \emph{arXiv preprint arXiv:2306.16928}, 2023{\natexlab{a}}.

\bibitem[Liu et~al.(2023{\natexlab{b}})Liu, Wu, Van~Hoorick, Tokmakov, Zakharov, and Vondrick]{liu2023zero}
Ruoshi Liu, Rundi Wu, Basile Van~Hoorick, Pavel Tokmakov, Sergey Zakharov, and Carl Vondrick.
\newblock Zero-1-to-3: Zero-shot one image to 3d object.
\newblock In \emph{Proceedings of the IEEE/CVF International Conference on Computer Vision}, pages 9298--9309, 2023{\natexlab{b}}.

\bibitem[Liu et~al.(2023{\natexlab{c}})Liu, Lin, Zeng, Long, Liu, Komura, and Wang]{liu2023syncdreamer}
Yuan Liu, Cheng Lin, Zijiao Zeng, Xiaoxiao Long, Lingjie Liu, Taku Komura, and Wenping Wang.
\newblock Syncdreamer: Generating multiview-consistent images from a single-view image.
\newblock \emph{arXiv preprint arXiv:2309.03453}, 2023{\natexlab{c}}.

\bibitem[Long et~al.(2022)Long, Lin, Wang, Komura, and Wang]{long2022sparseneus}
Xiaoxiao Long, Cheng Lin, Peng Wang, Taku Komura, and Wenping Wang.
\newblock Sparseneus: Fast generalizable neural surface reconstruction from sparse views.
\newblock In \emph{European Conference on Computer Vision}, pages 210--227. Springer, 2022.

\bibitem[Long et~al.(2023)Long, Guo, Lin, Liu, Dou, Liu, Ma, Zhang, Habermann, Theobalt, et~al.]{long2023wonder3d}
Xiaoxiao Long, Yuan-Chen Guo, Cheng Lin, Yuan Liu, Zhiyang Dou, Lingjie Liu, Yuexin Ma, Song-Hai Zhang, Marc Habermann, Christian Theobalt, et~al.
\newblock Wonder3d: Single image to 3d using cross-domain diffusion.
\newblock \emph{arXiv preprint arXiv:2310.15008}, 2023.

\bibitem[Metzer et~al.(2023)Metzer, Richardson, Patashnik, Giryes, and Cohen-Or]{metzer2023latent}
Gal Metzer, Elad Richardson, Or Patashnik, Raja Giryes, and Daniel Cohen-Or.
\newblock Latent-nerf for shape-guided generation of 3d shapes and textures.
\newblock In \emph{Proceedings of the IEEE/CVF Conference on Computer Vision and Pattern Recognition}, pages 12663--12673, 2023.

\bibitem[Michel et~al.(2022)Michel, Bar-On, Liu, Benaim, and Hanocka]{michel2022text2mesh}
Oscar Michel, Roi Bar-On, Richard Liu, Sagie Benaim, and Rana Hanocka.
\newblock Text2mesh: Text-driven neural stylization for meshes.
\newblock In \emph{Proceedings of the IEEE/CVF Conference on Computer Vision and Pattern Recognition}, pages 13492--13502, 2022.

\bibitem[Mohammad~Khalid et~al.(2022)Mohammad~Khalid, Xie, Belilovsky, and Popa]{mohammad2022clip}
Nasir Mohammad~Khalid, Tianhao Xie, Eugene Belilovsky, and Tiberiu Popa.
\newblock Clip-mesh: Generating textured meshes from text using pretrained image-text models.
\newblock In \emph{SIGGRAPH Asia 2022 conference papers}, pages 1--8, 2022.

\bibitem[Munkberg et~al.(2022)Munkberg, Hasselgren, Shen, Gao, Chen, Evans, M{\"u}ller, and Fidler]{munkberg2022extracting}
Jacob Munkberg, Jon Hasselgren, Tianchang Shen, Jun Gao, Wenzheng Chen, Alex Evans, Thomas M{\"u}ller, and Sanja Fidler.
\newblock Extracting triangular 3d models, materials, and lighting from images.
\newblock In \emph{Proceedings of the IEEE/CVF Conference on Computer Vision and Pattern Recognition}, pages 8280--8290, 2022.

\bibitem[Nichol et~al.(2021)Nichol, Dhariwal, Ramesh, Shyam, Mishkin, McGrew, Sutskever, and Chen]{nichol2021glide}
Alex Nichol, Prafulla Dhariwal, Aditya Ramesh, Pranav Shyam, Pamela Mishkin, Bob McGrew, Ilya Sutskever, and Mark Chen.
\newblock Glide: Towards photorealistic image generation and editing with text-guided diffusion models.
\newblock \emph{arXiv preprint arXiv:2112.10741}, 2021.

\bibitem[Podell et~al.(2023)Podell, English, Lacey, Blattmann, Dockhorn, M{\"u}ller, Penna, and Rombach]{podell2023sdxl}
Dustin Podell, Zion English, Kyle Lacey, Andreas Blattmann, Tim Dockhorn, Jonas M{\"u}ller, Joe Penna, and Robin Rombach.
\newblock Sdxl: Improving latent diffusion models for high-resolution image synthesis.
\newblock \emph{arXiv preprint arXiv:2307.01952}, 2023.

\bibitem[Poole et~al.(2022)Poole, Jain, Barron, and Mildenhall]{poole2022dreamfusion}
Ben Poole, Ajay Jain, Jonathan~T Barron, and Ben Mildenhall.
\newblock Dreamfusion: Text-to-3d using 2d diffusion.
\newblock \emph{arXiv preprint arXiv:2209.14988}, 2022.

\bibitem[Qian et~al.(2023)Qian, Mai, Hamdi, Ren, Siarohin, Li, Lee, Skorokhodov, Wonka, Tulyakov, et~al.]{qian2023magic123}
Guocheng Qian, Jinjie Mai, Abdullah Hamdi, Jian Ren, Aliaksandr Siarohin, Bing Li, Hsin-Ying Lee, Ivan Skorokhodov, Peter Wonka, Sergey Tulyakov, et~al.
\newblock Magic123: One image to high-quality 3d object generation using both 2d and 3d diffusion priors.
\newblock \emph{arXiv preprint arXiv:2306.17843}, 2023.

\bibitem[Radford et~al.(2021)Radford, Kim, Hallacy, Ramesh, Goh, Agarwal, Sastry, Askell, Mishkin, Clark, et~al.]{radford2021learning}
Alec Radford, Jong~Wook Kim, Chris Hallacy, Aditya Ramesh, Gabriel Goh, Sandhini Agarwal, Girish Sastry, Amanda Askell, Pamela Mishkin, Jack Clark, et~al.
\newblock Learning transferable visual models from natural language supervision.
\newblock In \emph{International conference on machine learning}, pages 8748--8763. PMLR, 2021.

\bibitem[Ramesh et~al.(2022)Ramesh, Dhariwal, Nichol, Chu, and Chen]{ramesh2022hierarchical}
Aditya Ramesh, Prafulla Dhariwal, Alex Nichol, Casey Chu, and Mark Chen.
\newblock Hierarchical text-conditional image generation with clip latents.
\newblock \emph{arXiv preprint arXiv:2204.06125}, 1\penalty0 (2):\penalty0 3, 2022.

\bibitem[Richardson et~al.(2023)Richardson, Metzer, Alaluf, Giryes, and Cohen-Or]{richardson2023texture}
Elad Richardson, Gal Metzer, Yuval Alaluf, Raja Giryes, and Daniel Cohen-Or.
\newblock Texture: Text-guided texturing of 3d shapes.
\newblock \emph{arXiv preprint arXiv:2302.01721}, 2023.

\bibitem[Rombach et~al.(2022)Rombach, Blattmann, Lorenz, Esser, and Ommer]{rombach2022high}
Robin Rombach, Andreas Blattmann, Dominik Lorenz, Patrick Esser, and Bj{\"o}rn Ommer.
\newblock High-resolution image synthesis with latent diffusion models.
\newblock In \emph{Proceedings of the IEEE/CVF conference on computer vision and pattern recognition}, pages 10684--10695, 2022.

\bibitem[Saharia et~al.(2022)Saharia, Chan, Saxena, Li, Whang, Denton, Ghasemipour, Gontijo~Lopes, Karagol~Ayan, Salimans, et~al.]{saharia2022photorealistic}
Chitwan Saharia, William Chan, Saurabh Saxena, Lala Li, Jay Whang, Emily~L Denton, Kamyar Ghasemipour, Raphael Gontijo~Lopes, Burcu Karagol~Ayan, Tim Salimans, et~al.
\newblock Photorealistic text-to-image diffusion models with deep language understanding.
\newblock \emph{Advances in Neural Information Processing Systems}, 35:\penalty0 36479--36494, 2022.

\bibitem[Schuhmann et~al.(2022)Schuhmann, Beaumont, Vencu, Gordon, Wightman, Cherti, Coombes, Katta, Mullis, Wortsman, et~al.]{schuhmann2022laion}
Christoph Schuhmann, Romain Beaumont, Richard Vencu, Cade Gordon, Ross Wightman, Mehdi Cherti, Theo Coombes, Aarush Katta, Clayton Mullis, Mitchell Wortsman, et~al.
\newblock Laion-5b: An open large-scale dataset for training next generation image-text models.
\newblock \emph{Advances in Neural Information Processing Systems}, 35:\penalty0 25278--25294, 2022.

\bibitem[Shi et~al.(2023)Shi, Wang, Ye, Long, Li, and Yang]{shi2023mvdream}
Yichun Shi, Peng Wang, Jianglong Ye, Mai Long, Kejie Li, and Xiao Yang.
\newblock Mvdream: Multi-view diffusion for 3d generation.
\newblock \emph{arXiv preprint arXiv:2308.16512}, 2023.

\bibitem[Siddiqui et~al.(2022)Siddiqui, Thies, Ma, Shan, Nie{\ss}ner, and Dai]{siddiqui2022texturify}
Yawar Siddiqui, Justus Thies, Fangchang Ma, Qi Shan, Matthias Nie{\ss}ner, and Angela Dai.
\newblock Texturify: Generating textures on 3d shape surfaces.
\newblock In \emph{European Conference on Computer Vision}, pages 72--88. Springer, 2022.

\bibitem[Song et~al.(2020)Song, Meng, and Ermon]{song2020denoising}
Jiaming Song, Chenlin Meng, and Stefano Ermon.
\newblock Denoising diffusion implicit models.
\newblock \emph{arXiv preprint arXiv:2010.02502}, 2020.

\bibitem[Sun et~al.(2024)Sun, Wu, and Gao]{Sun2024}
Jia-Mu Sun, Tong Wu, and Lin Gao.
\newblock Recent advances in implicit representation-based 3d shape generation.
\newblock \emph{Visual Intelligence}, 2\penalty0 (1):\penalty0 9, 2024.

\bibitem[Szegedy et~al.(2016)Szegedy, Vanhoucke, Ioffe, Shlens, and Wojna]{szegedy2016rethinking}
Christian Szegedy, Vincent Vanhoucke, Sergey Ioffe, Jon Shlens, and Zbigniew Wojna.
\newblock Rethinking the inception architecture for computer vision.
\newblock In \emph{Proceedings of the IEEE conference on computer vision and pattern recognition}, pages 2818--2826, 2016.

\bibitem[Szymanowicz et~al.(2023)Szymanowicz, Rupprecht, and Vedaldi]{szymanowicz2023viewset}
Stanislaw Szymanowicz, Christian Rupprecht, and Andrea Vedaldi.
\newblock Viewset diffusion:(0-) image-conditioned 3d generative models from 2d data.
\newblock \emph{arXiv preprint arXiv:2306.07881}, 2023.

\bibitem[Tang et~al.(2023)Tang, Zhang, Chen, Wang, and Furukawa]{tang2023mvdiffusion}
Shitao Tang, Fuyang Zhang, Jiacheng Chen, Peng Wang, and Yasutaka Furukawa.
\newblock Mvdiffusion: Enabling holistic multi-view image generation with correspondence-aware diffusion.
\newblock \emph{arXiv preprint arXiv:2307.01097}, 2023.

\bibitem[Tseng et~al.(2023)Tseng, Li, Kim, Alsisan, Huang, and Kopf]{tseng2023consistent}
Hung-Yu Tseng, Qinbo Li, Changil Kim, Suhib Alsisan, Jia-Bin Huang, and Johannes Kopf.
\newblock Consistent view synthesis with pose-guided diffusion models.
\newblock In \emph{Proceedings of the IEEE/CVF Conference on Computer Vision and Pattern Recognition}, pages 16773--16783, 2023.

\bibitem[Turk and Levoy(1994)]{turk1994zippered}
Greg Turk and Marc Levoy.
\newblock Zippered polygon meshes from range images.
\newblock In \emph{Proceedings of the 21st annual conference on Computer graphics and interactive techniques}, pages 311--318, 1994.

\bibitem[Wang et~al.(2023{\natexlab{a}})Wang, Du, Li, Yeh, and Shakhnarovich]{wang2023score}
Haochen Wang, Xiaodan Du, Jiahao Li, Raymond~A Yeh, and Greg Shakhnarovich.
\newblock Score jacobian chaining: Lifting pretrained 2d diffusion models for 3d generation.
\newblock In \emph{Proceedings of the IEEE/CVF Conference on Computer Vision and Pattern Recognition}, pages 12619--12629, 2023{\natexlab{a}}.

\bibitem[Wang et~al.(2023{\natexlab{b}})Wang, Lu, Wang, Bao, Li, Su, and Zhu]{wang2023prolificdreamer}
Zhengyi Wang, Cheng Lu, Yikai Wang, Fan Bao, Chongxuan Li, Hang Su, and Jun Zhu.
\newblock Prolificdreamer: High-fidelity and diverse text-to-3d generation with variational score distillation.
\newblock \emph{arXiv preprint arXiv:2305.16213}, 2023{\natexlab{b}}.

\bibitem[Xiang et~al.(2023)Xiang, Yang, Huang, and Tong]{xiang20233d}
Jianfeng Xiang, Jiaolong Yang, Binbin Huang, and Xin Tong.
\newblock 3d-aware image generation using 2d diffusion models.
\newblock \emph{arXiv preprint arXiv:2303.17905}, 2023.

\bibitem[Yang et~al.(2023)Yang, Zhou, Liu, and Loy]{yang2023rerender}
Shuai Yang, Yifan Zhou, Ziwei Liu, and Chen~Change Loy.
\newblock Rerender a video: Zero-shot text-guided video-to-video translation.
\newblock \emph{arXiv preprint arXiv:2306.07954}, 2023.

\bibitem[Young(2016)]{xatlas2016}
Jonathan Young.
\newblock xatlas.
\newblock In \emph{github.com/jpcy/xatlas}, 2016.

\bibitem[Zhang et~al.(2023)Zhang, Rao, and Agrawala]{zhang2023adding}
Lvmin Zhang, Anyi Rao, and Maneesh Agrawala.
\newblock Adding conditional control to text-to-image diffusion models.
\newblock In \emph{Proceedings of the IEEE/CVF International Conference on Computer Vision}, pages 3836--3847, 2023.

\end{thebibliography}
}

\clearpage
\setcounter{page}{1}
\maketitlesupplementary

\section{Implementation Details}
\begin{table}[b]
\centering
\begin{tabular}{cc|cc} 
\toprule
\multicolumn{4}{c}{$\mathcal{C}^{(sampling)}$}\\
\midrule
elevation & azimuth & elevation & azimuth\\
$0^\circ$ & $0^\circ$  & $0^\circ$ & $180^\circ$ \\
$0^\circ$ & $90^\circ$ & $0^\circ$ & $270^\circ$ \\
\midrule
\multicolumn{4}{c}{$\mathcal{C}^{(inpainting)}$}\\
\midrule
elevation & azimuth & elevation & azimuth\\
$90^\circ$ & $0^\circ$ & $60^\circ$ & $315^\circ$\\
$0^\circ$ & $45^\circ$ & $60^\circ$ & $90^\circ$\\
$0^\circ$ & $315^\circ$ & $60^\circ$ & $270^\circ$\\
$0^\circ$ & $135^\circ$ & $60^\circ$ & $135^\circ$\\
$0^\circ$ & $225^\circ$ & $60^\circ$ & $225^\circ$\\
$60^\circ$ & $0^\circ$ & $60^\circ$ & $180^\circ$\\
$60^\circ$ & $45^\circ$ & & \\
\midrule
\multicolumn{4}{c}{$\mathcal{C}^{(img2img)}$}\\
\midrule
elevation & azimuth & elevation & azimuth\\
$0^\circ$ & $180^\circ$ & $0^\circ$ & $270^\circ$\\
$0^\circ$ & $135^\circ$ & $0^\circ$ & $45^\circ$\\
$0^\circ$ & $225^\circ$ & $0^\circ$ & $315^\circ$\\
$0^\circ$ & $90^\circ$ & $0^\circ$ & $0^\circ$\\
\bottomrule
\end{tabular}
\caption{ \small Viewpoints settings.
}
\label{tab: camera}
\end{table}
\textbf{Detailed Parameters.} We set 3 sets of camera viewpoints $\mathcal{C}^{(sampling)}$, $\mathcal{C}^{(inpainting)}$ and $\mathcal{C}^{(img2img)}$ for texture space samling, Inpainting epoch and Img2Img epoch, respectively. We use the same viewpoints configuration for all the inputs, as shown in Tab.~\ref{tab: camera}. We disable `guess mode', \ie, we did not apply depth control to the unconditional guidance side of the classifier-free guidance because guess mode tends to produce unnatural colors. Following the original DDIM~\cite{song2020denoising}, in the denoising process we set  $\sigma_i = \sqrt{(1-\alpha_{i-1})/(1-\alpha_i)} \sqrt{1 - \alpha_i / \alpha_{i-1}}$ and use a linear time schedule $\{t_i\}_{i=T}^0$.

\begin{figure}[tb]
  \centering
  \includegraphics[width=\linewidth]{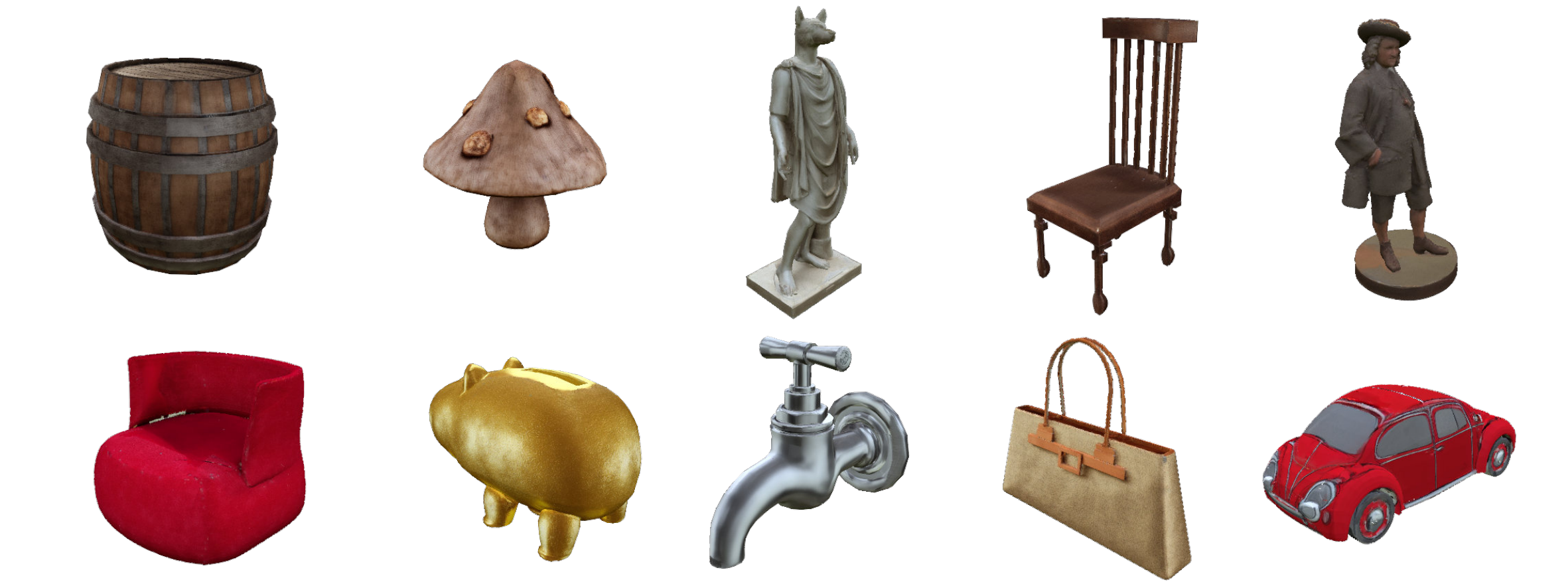}
  \caption{
    Renderings with `\textit{diffuse}' BSDF.
  }\label{fig:supplight}
\end{figure}

\noindent \textbf{Rendering Settings.} We modify nvdiffrec~\cite{munkberg2022extracting}, which is based on the differentiable rendering pipeline implemented using nvdiffrast~\cite{laine2020modular}, to implement the rendering function $\mathcal{R}$. We set the BSDF (bidirectional scattering distribution function) type to `$k_d$' to ignore the influence of lighting, as in this work, we focus on generating the content of textures rather than decoupling materials from lighting. The visual results presented in the main paper also use `$k_d$' as the BSDF type. In Fig.~\ref{fig:supplight}, we show the rendering results with `\textit{diffuse}' as the BSDF type, using a museumplein environment light. It can be observed that some inconsistent light-dark relationships appear in `\textit{diffuse}' rendering results. We will explore generating texture maps that comply with the Physically Based Rendering (PBR) workflow in future work.
\begin{figure}[tb]
  \centering
  \includegraphics[width=\linewidth]{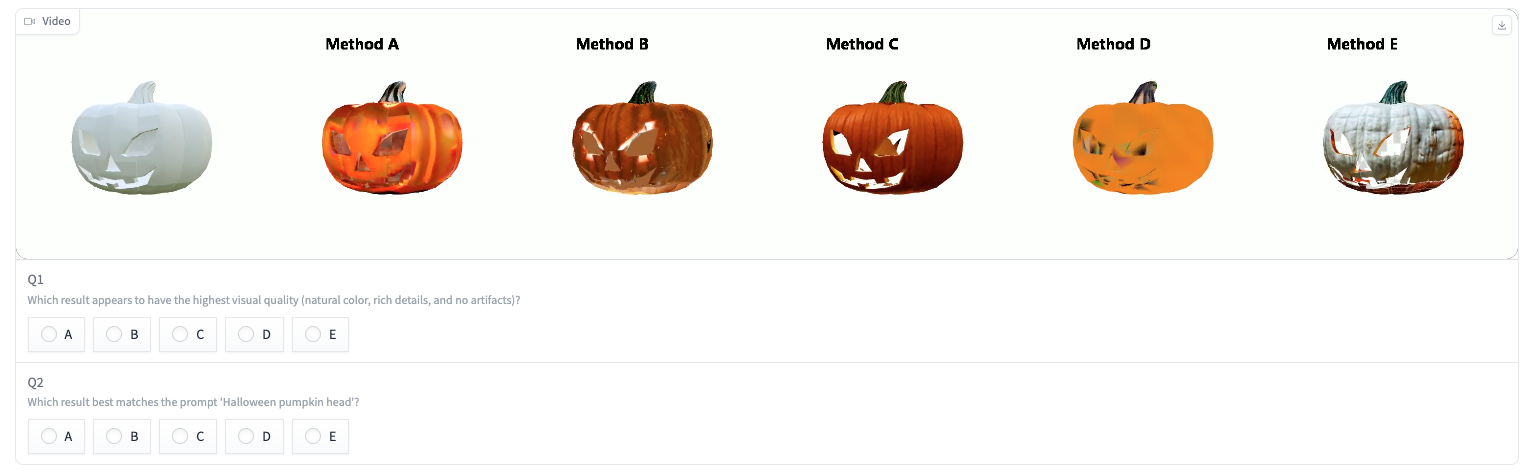}
  \caption{
    Screenshot of user study.
  }\label{fig:suppuser}
\end{figure}

\noindent \textbf{User Study.} 
To compare with the baseline methods, we conduct a user study as part of the evaluation. We implement a survey using Gradio~\cite{abid2019gradio}, which is a webpage-based tool. The survey randomly present 10 groups of generated results to each participant. A screenshot of the survey for a group of generated results is displayed in Fig.~\ref{fig:suppuser}, which includes six videos and two questions:
\begin{enumerate}
    \item \textit{Which result appears to have the highest visual quality (natural color, rich details, and no artifacts)?}
    \item \textit{Which result best matches the prompt `[prompt]'?}
\end{enumerate}
For each group of results displayed in the videos, we ensure that their order is randomly shuffled to prevent bias. Responses where all answers have the same selection and responses with completely identical answers are considered invalid. After filtering, we obtain a total of 35 valid surveys.

\begin{table}[t]
\small
\centering
\begin{tabular}{|c|c|c|c|} 
\hline
\diagbox{Loss}{Win} & Ours & Text2Tex & TEXTure \\
 \hline
Ours   & - & 20 & 24 \\
\hline
Text2Tex  & 60 & - & 43 \\
\hline
TEXTure & 56 & 37 & - \\
\hline
\end{tabular}
\caption{\small Results of pairwise user study.
}
\label{tab: pairwise}
\end{table}

We also conduct a pairwise comparison test with two competitive methods, as shown in Tab.~\ref{tab: pairwise}. We employ the Bradley-Terry model to analyze the results of the pairwise user study. The estimated Bradley-Terry model parameters $p_\texttt{Ours}, p_\texttt{Text2Tex}, p_\texttt{TEXTure}$ are $1.91, 0.66, 0.79$ respectively, which indicates that ours is the strongest.

\section{More Results}
\textbf{We highly recommend readers to visit our project homepage\footnote{\url{https://cjeen.github.io/GenesisTexPaper/}} to view the result videos.}
\subsection{Comparison Results}
\begin{figure*}[t]
  \centering
  \includegraphics[width=0.9\linewidth]{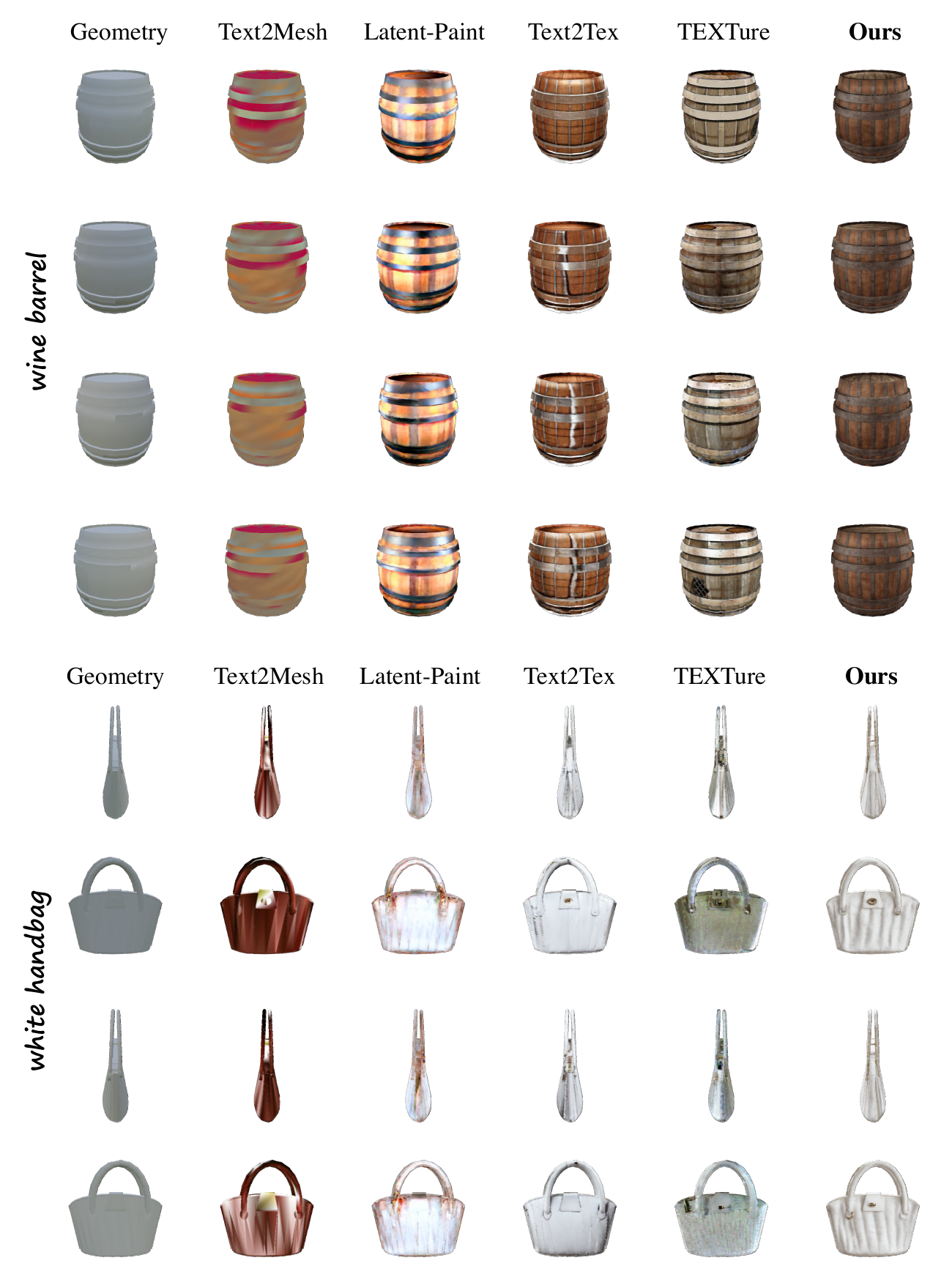}
  \caption{
    More qualitative comparisons I.
  }\label{fig:suppcomp1}
\end{figure*}

\begin{figure*}[t]
  \centering
  \includegraphics[width=0.9\linewidth]{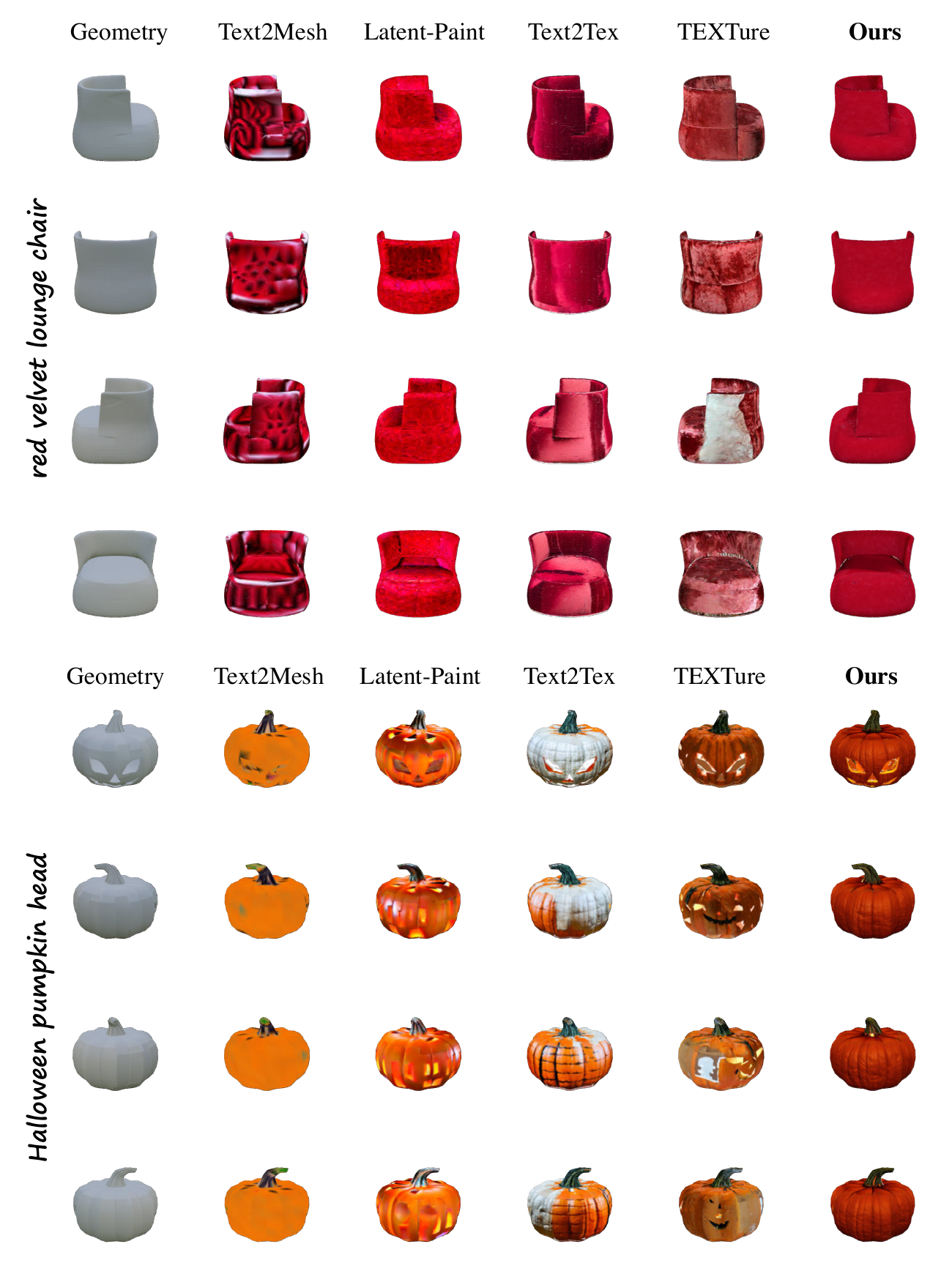}
  \caption{
    More qualitative comparisons II.
  }\label{fig:suppcomp2}
\end{figure*}

\begin{figure*}[t]
  \centering
  \includegraphics[width=0.9\linewidth]{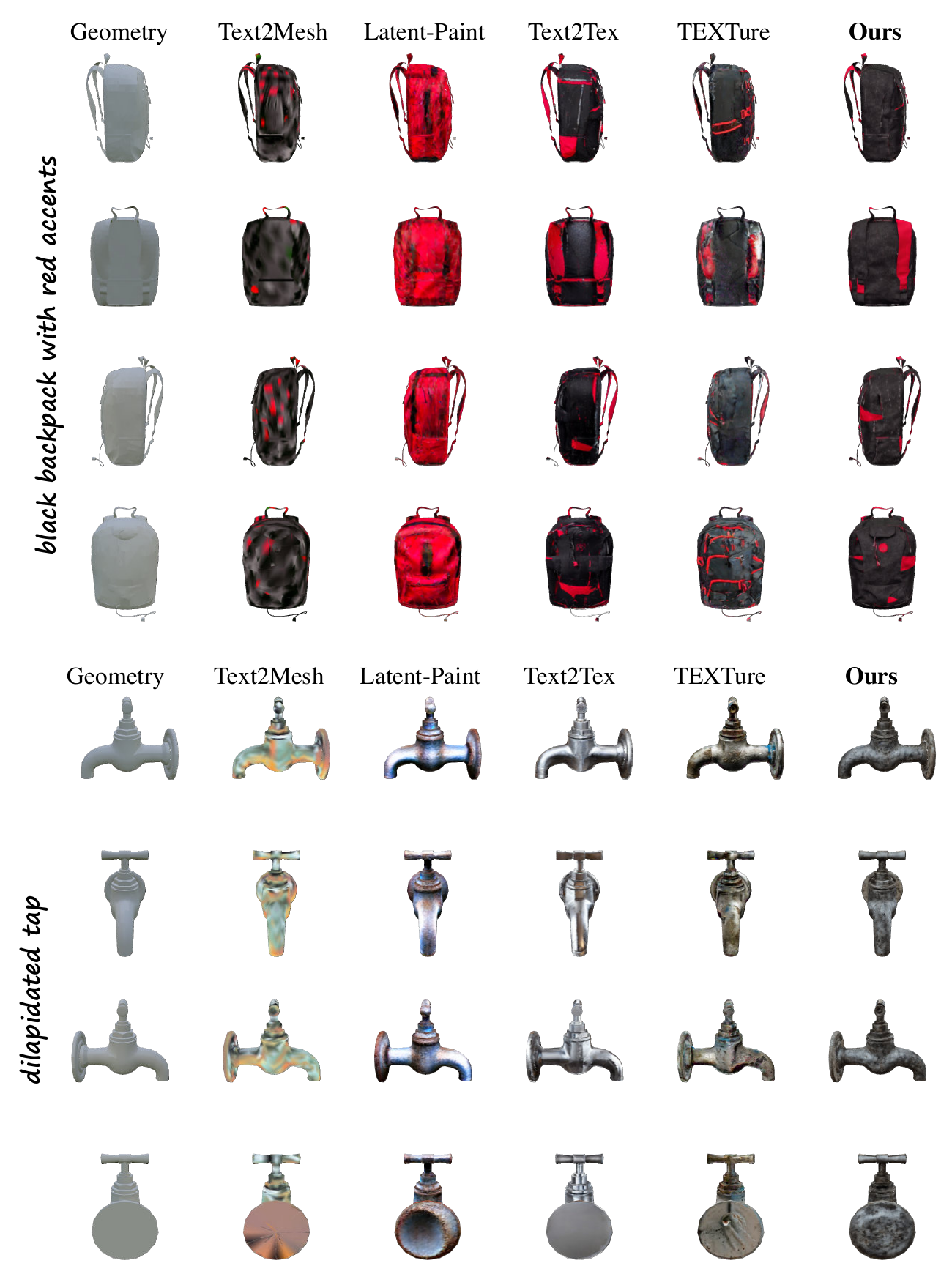}
  \caption{
    More qualitative comparisons III.
  }\label{fig:suppcomp3}
\end{figure*}

\begin{figure*}[t]
  \centering
  \includegraphics[width=0.9\linewidth]{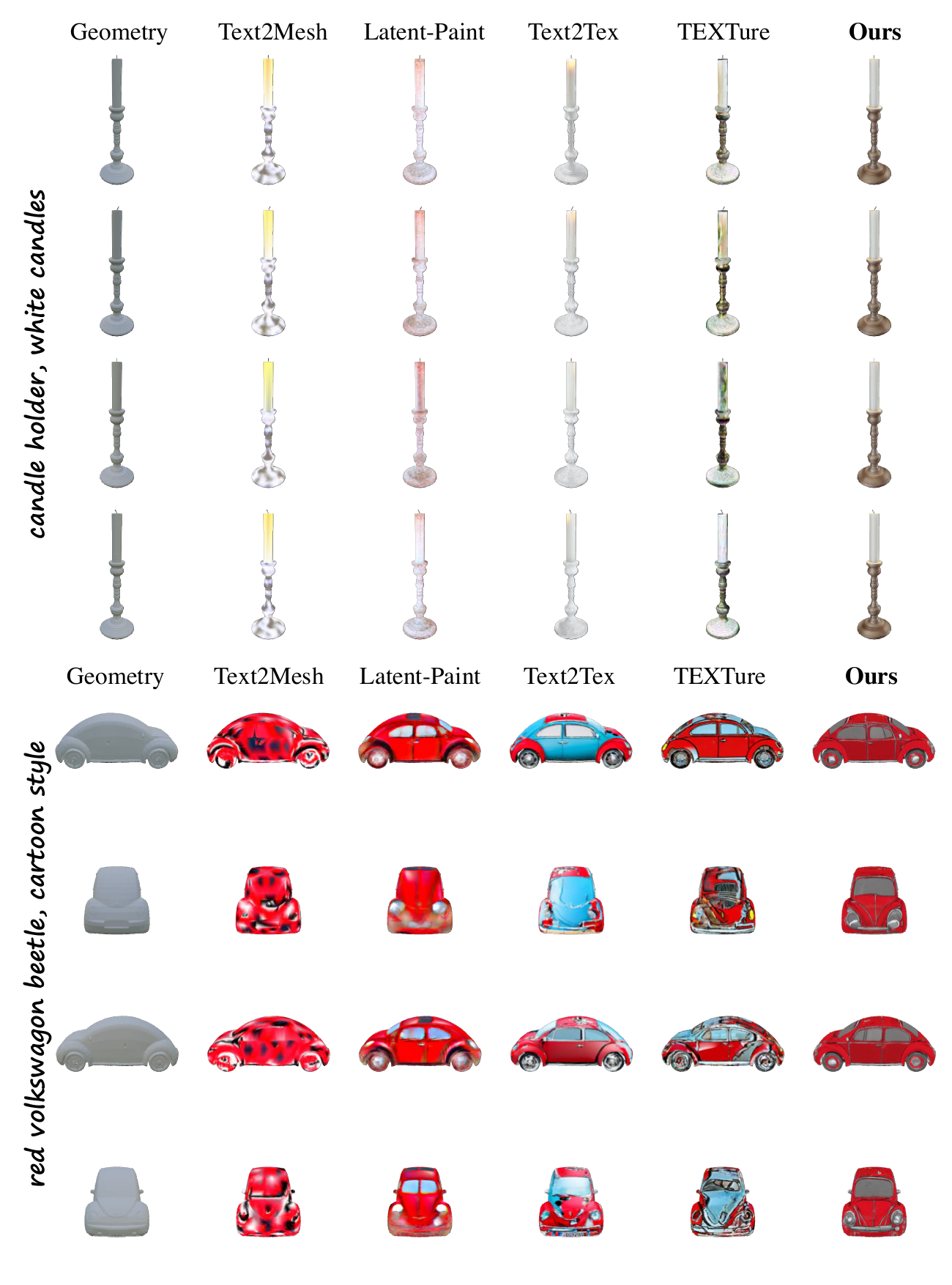}
  \caption{
    More qualitative comparisons IV.
  }\label{fig:suppcomp4}
\end{figure*}

\begin{figure*}[t]
  \centering
  \includegraphics[width=0.9\linewidth]{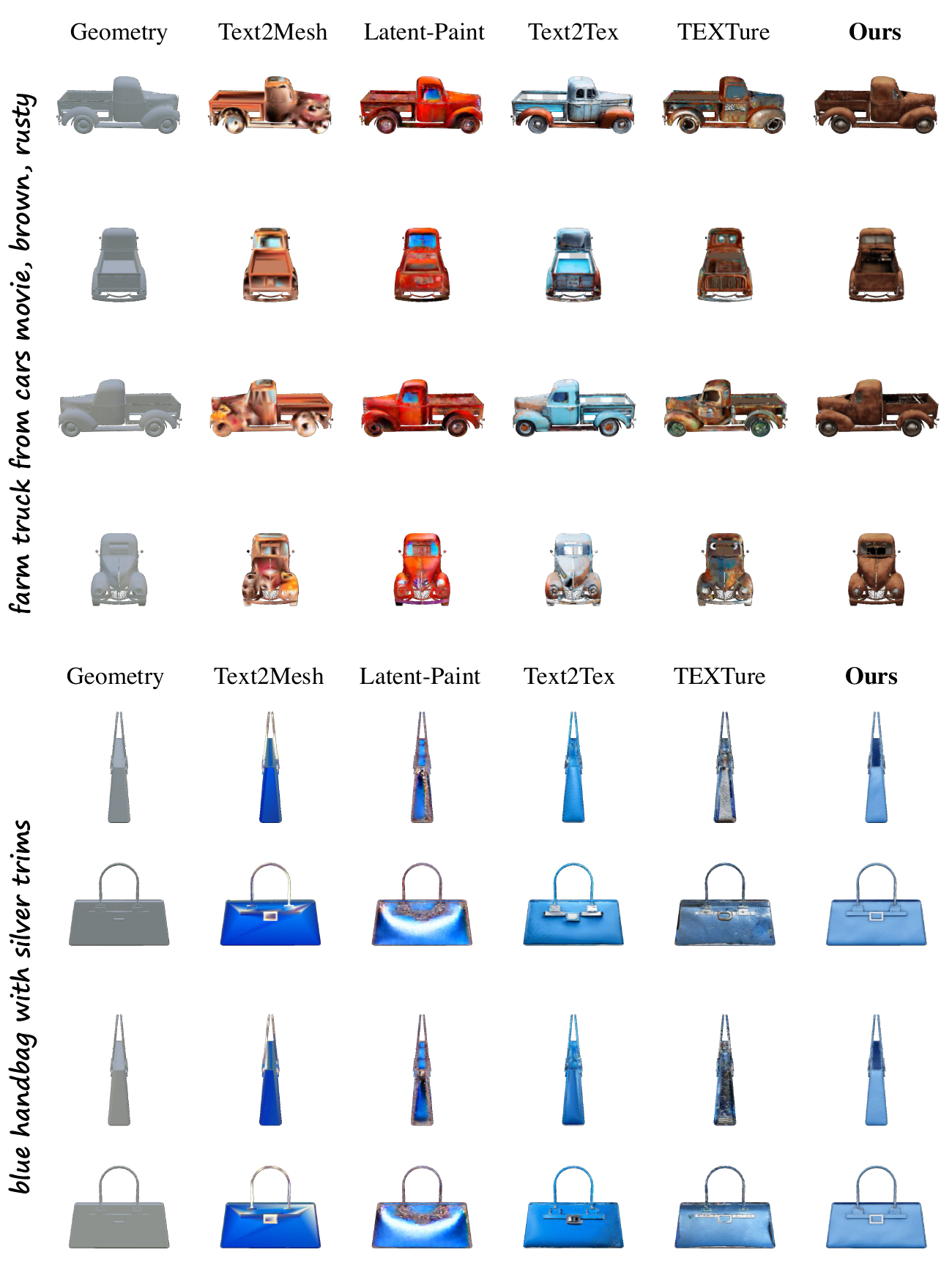}
  \caption{
    More qualitative comparisons V.
  }\label{fig:suppcomp5}
\end{figure*}

\begin{figure*}[t]
  \centering
  \includegraphics[width=0.9\linewidth]{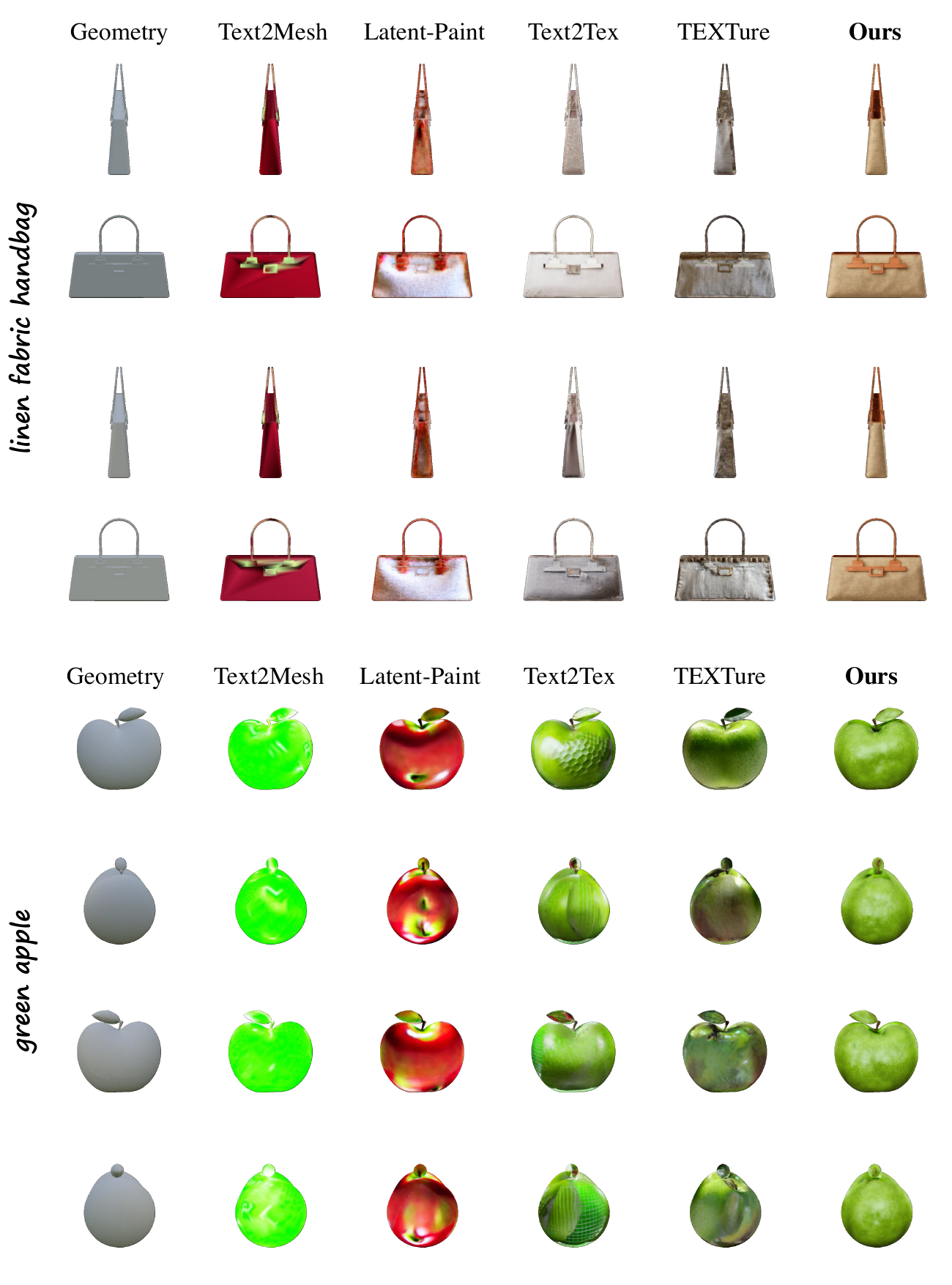}
  \caption{
    More qualitative comparisons VI.
  }\label{fig:suppcomp6}
\end{figure*}

We provide more visual comparisons between our method and state-of-the-art baselines~\cite{metzer2023latent, michel2022text2mesh, chen2023text2tex, richardson2023texture} in the video titled `\textit{Comparisons}'. In Fig.~\ref{fig:suppcomp1}-\ref{fig:suppcomp6}, we show some multi-view renderings from the video. It is clear from these comparisons that our method outperforms the baseline approaches in terms of both visual quality and alignment with the input prompt.
\subsection{Ablation Results}
\begin{figure*}[t]
  \centering
  \includegraphics[width=0.7\linewidth]{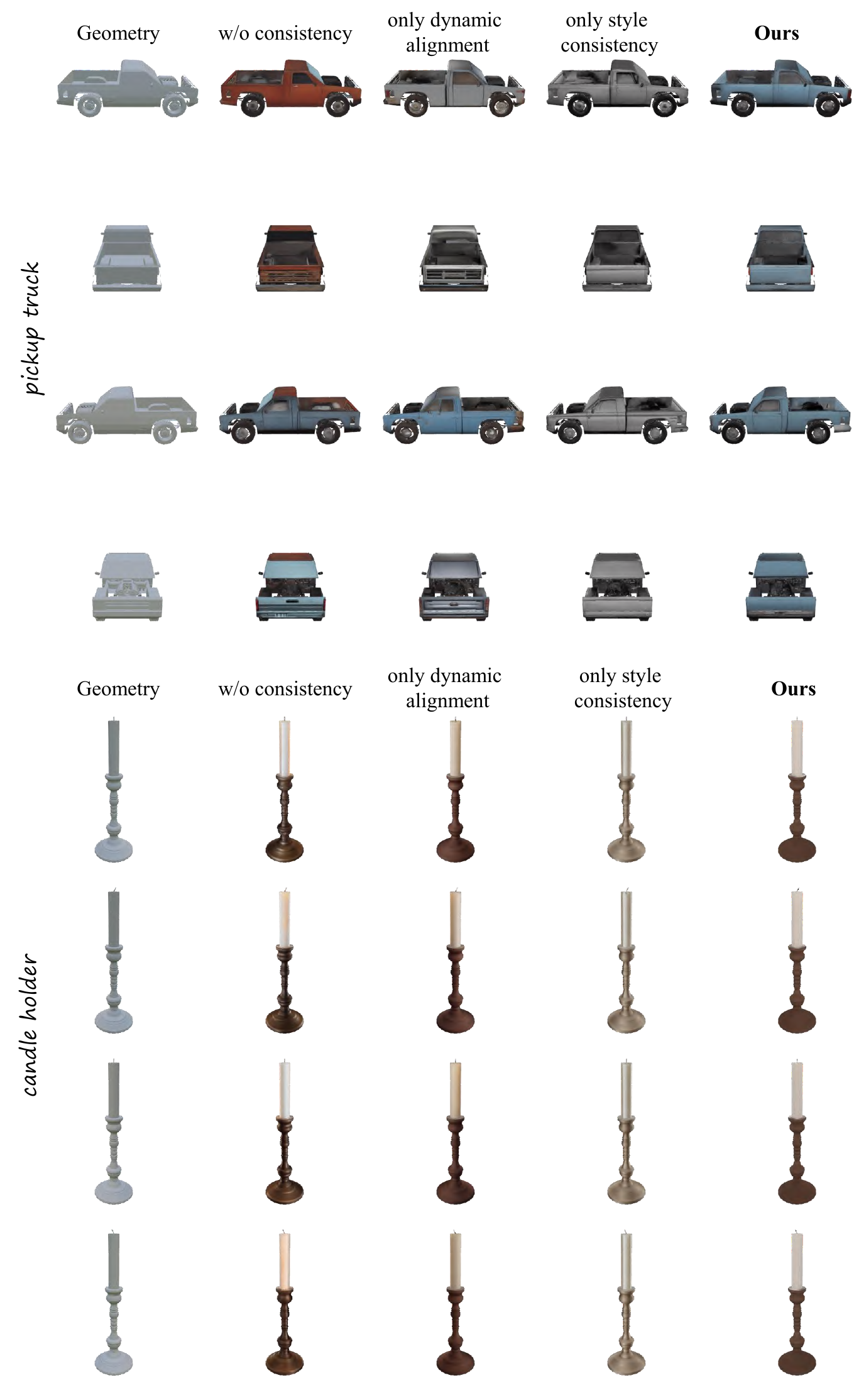}
  \caption{
    More ablation results I.
  }\label{fig:suppabl1}
\end{figure*}

\begin{figure*}[t]
  \centering
  \includegraphics[width=0.7\linewidth]{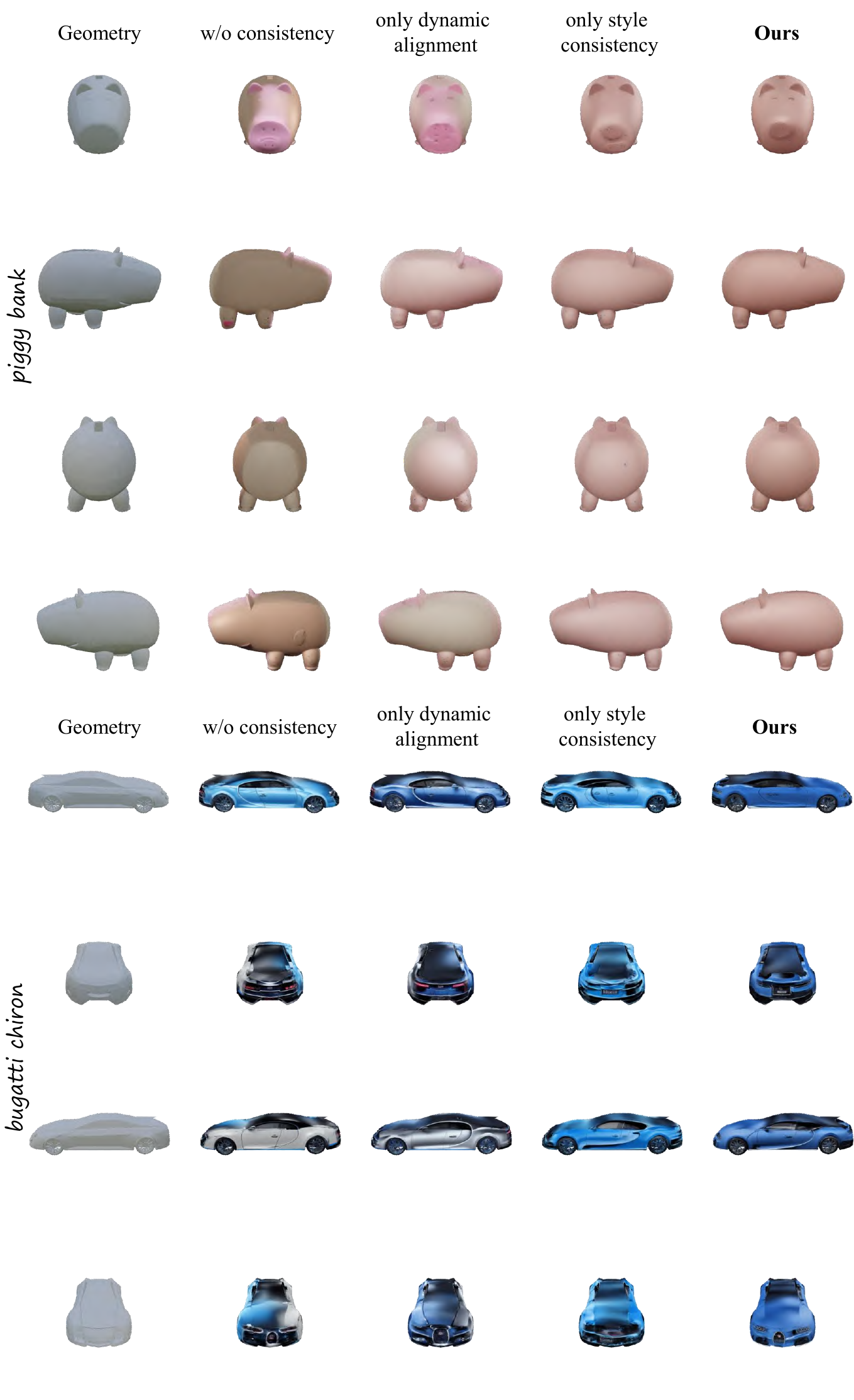}
  \caption{
    More ablation results II.
  }\label{fig:suppabl2}
\end{figure*}

In texture space sampling, we leverage dynamic alignment and style consistency to ensure consistency across multiple viewpoints. To verify the effectiveness of these two operations on the results, we present a visual comparison of the generated results under different consistency settings in the video titled `\textit{Consistency Ablations}'. In Fig.~\ref{fig:suppabl1}-\ref{fig:suppabl2}, we show some multi-view renderings from the video. It can be observed that style consistency greatly affects the global style harmony, while dynamic alignment can resolve multi-view conflicts.
\subsection{Stable Diffusion XL Generation Results}
\begin{figure*}[t]
  \centering
  \includegraphics[width=0.9\linewidth]{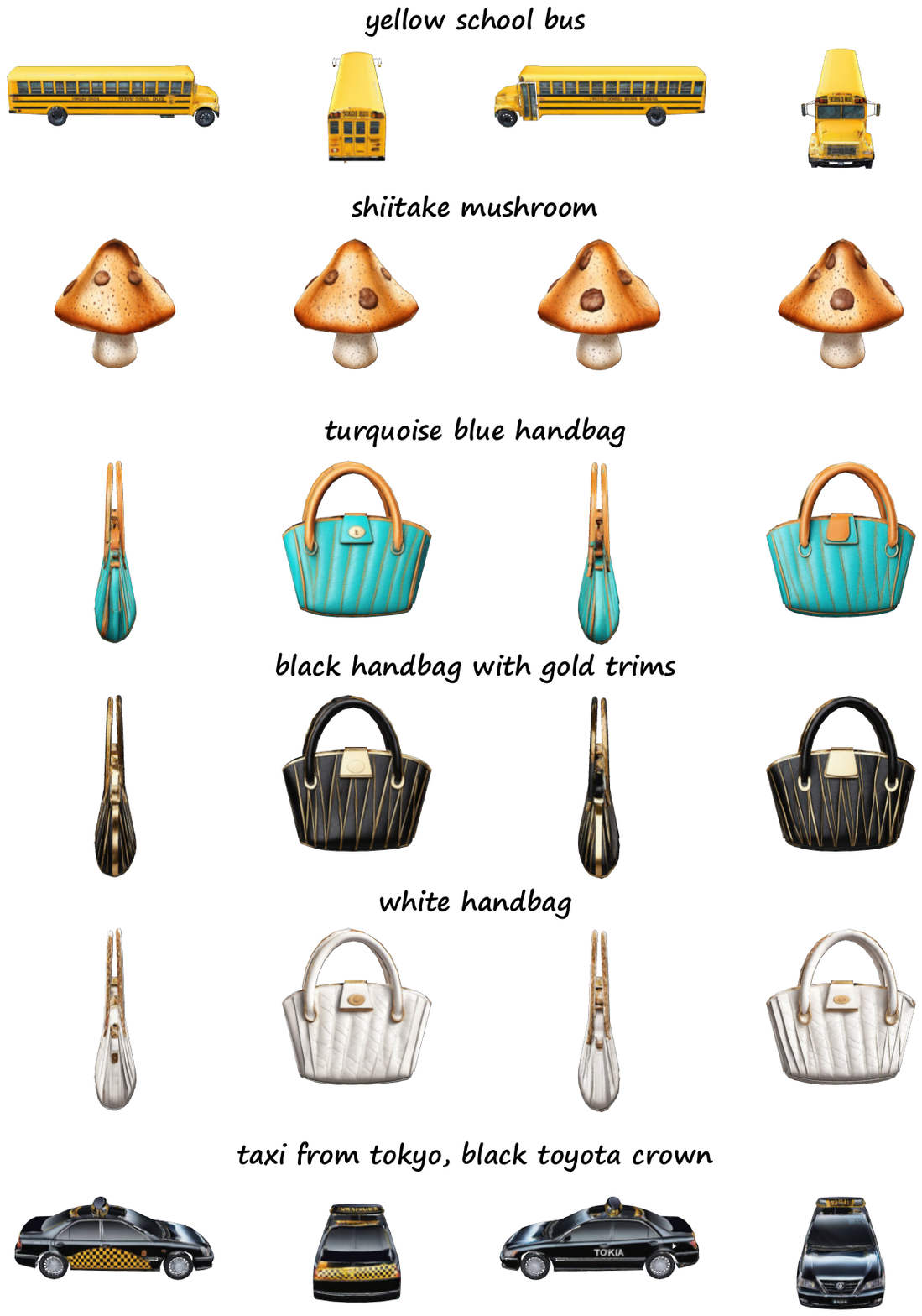}
  \caption{
    Generation results with Stable Diffusion XL I.
  }\label{fig:suppxl1}
\end{figure*}

\begin{figure*}[t]
  \centering
  \includegraphics[width=0.9\linewidth]{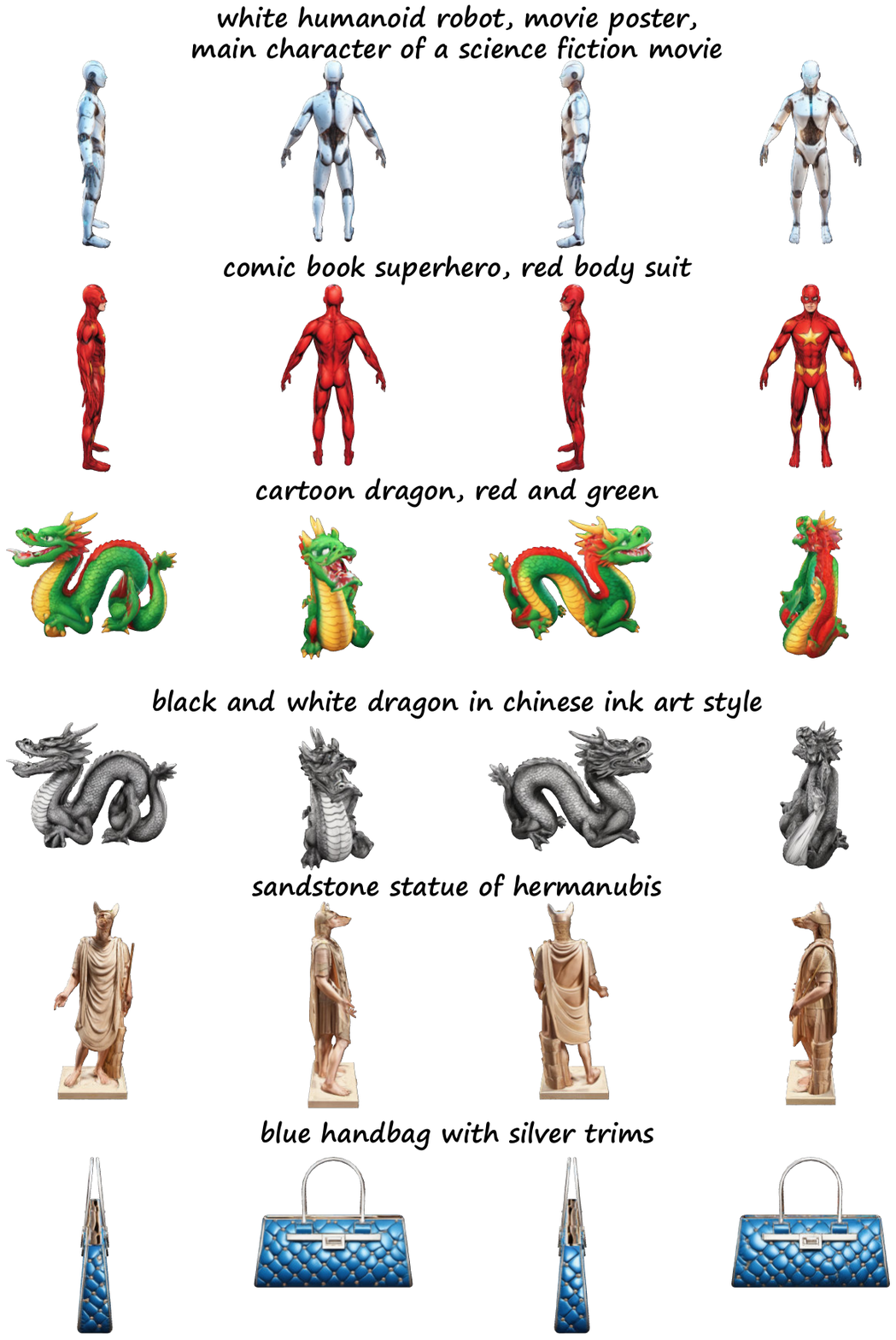}
  \caption{
    Generation results with Stable Diffusion XL II.
  }\label{fig:suppxl2}
\end{figure*}

In the main paper, we utilized \texttt{Stable Diffusion v1.5}~\cite{rombach2022high} as the image diffusion model. To further enhance the quality of generated textures, we conducted an experiment to explore the effectiveness of GenesisTex using \texttt{Stable Diffusion XL}~\cite{podell2023sdxl} for texture synthesis. The results are showcased in the video titled `\textit{Texturing with Stable Diffusion XL}'. Figure~\ref{fig:suppxl1} and Fig.~\ref{fig:suppxl2} displays some multi-view rendering images extracted from the video. It can be observed that our method, leveraging Stable Diffusion XL, produces textures with remarkably high detail quality and minimal artifacts. This experiment highlights the potential of our approach when applied with more powerful image diffusion models.




\end{document}